\documentclass{article}

\PassOptionsToPackage{numbers, compress}{natbib}

\usepackage[final]{neurips_2021} 

\usepackage[utf8]{inputenc} 
\usepackage[T1]{fontenc}    
\usepackage{url}            
\usepackage{booktabs}       
\usepackage{amsfonts}       
\usepackage{nicefrac}       
\usepackage{microtype}      

\usepackage{graphicx}
\usepackage[section]{placeins}
\usepackage{algorithm} 
\usepackage{algpseudocode}
\usepackage{subcaption}
\usepackage{amsmath}
\usepackage{svg}
\usepackage{comment}
\usepackage{tabularx}

\usepackage{xcolor}         
\usepackage{hyperref}       

\title{The Power of Communication in a Distributed Multi-Agent System}

\author{%
  Philipp D. Siedler \\
  Independent Researcher \\
  London, UK \\ 
  \texttt{p.d.siedler@gmail.com} \\
}

\begin{document}

\maketitle

\begin{abstract}
  
  Single-Agent (SA) Reinforcement Learning systems have shown outstanding results on non-stationary problems. However, Multi-Agent Reinforcement Learning (MARL) can surpass SA systems generally and when scaling. Furthermore, MA systems can be super-powered by collaboration, which can happen through observing others, or a communication system used to share information between collaborators. Here, we developed a distributed MA learning mechanism with the ability to communicate based on decentralised partially observable Markov decision processes (Dec-POMDPs) and Graph Neural Networks (GNNs). Minimising the time and energy consumed by training Machine Learning models while improving performance can be achieved by collaborative MA mechanisms. We demonstrate this in a real-world scenario, an offshore wind farm, including a set of distributed wind turbines, where the objective is to maximise collective efficiency. Compared to a SA system, MA collaboration has shown significantly reduced training time and higher cumulative rewards in unseen and scaled scenarios.
  
  
\end{abstract}

\begin{figure}[h]
  \centering
  \includegraphics[width=\textwidth]{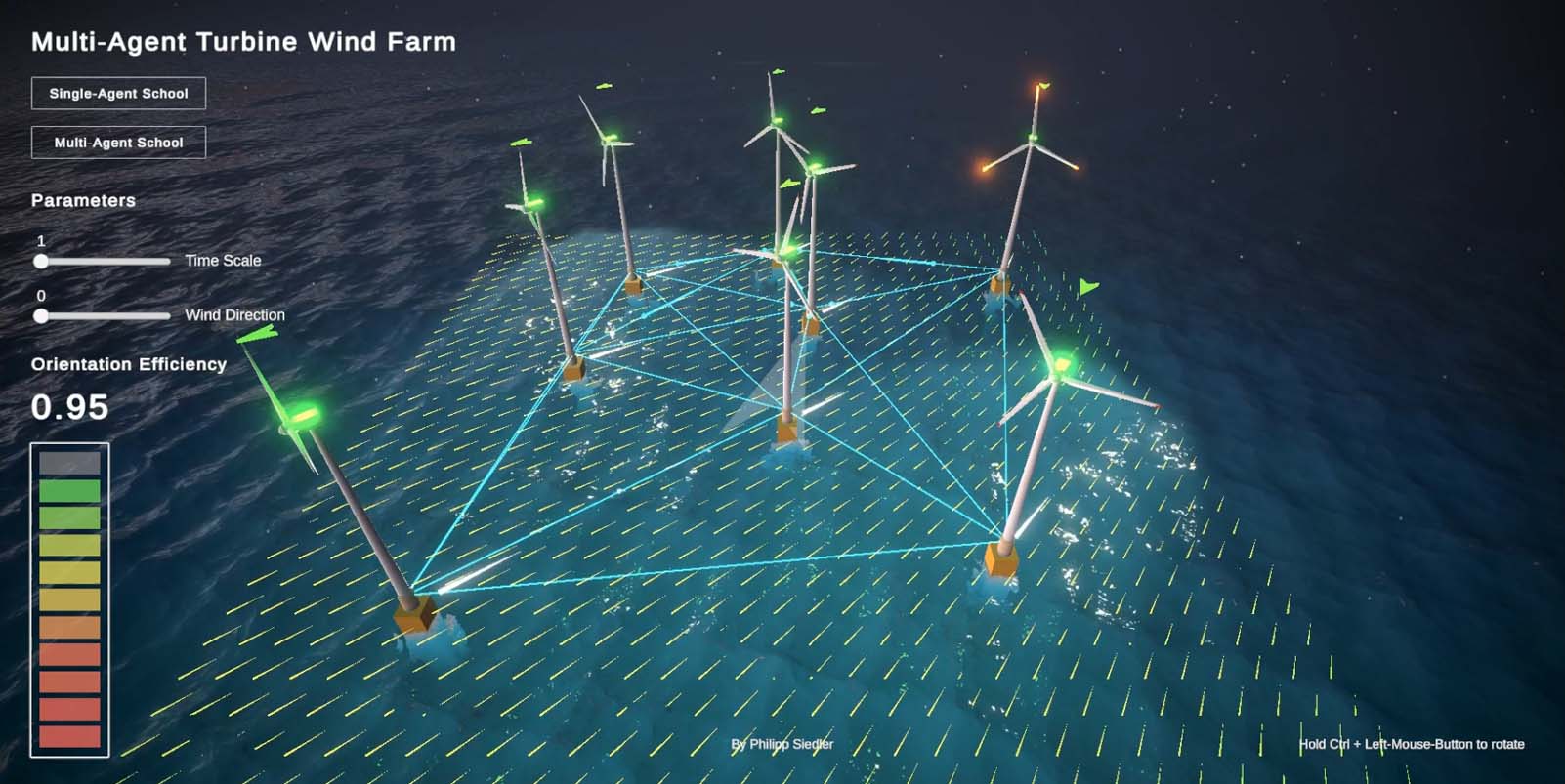}
  \caption{Dashboard of multi-agent wind farm environment in inference mode. Web-browser application can be found at: \url{https://philippds-pages.github.io/RL-Wind-Farm_WebApp/}.}
\end{figure}

\section{Introduction}
  \subsection{Motivation}
  In a Multi-Agent Reinforcement Learning (MARL) environment, groups of agents with baseline intelligence and ability can have a higher collective intelligence by behaving together \cite{cohen_team_1997}. A shared pool of information through a collective observation space can help individual agents to learn quicker, and as a group reach goals they might not have been able to reach on their own \cite{guestrin_coordinated_2002, decker_distributed_1987, panait_cooperative_2005, mataric_using_1998}. However, acting as a collective requires collaboration. From the perspective of an individual agent, other agents in the collective and the consequences of their actions, i.e. change of the environment, can be seen as part of a dynamic environment \cite{ravula_ad_2019}. Perceiving others actions and making sense of their intention is called intention reading, stated in the theory-of-mind (ToM) \cite{hernandez-leal_survey_2019-1}. While this is an integral part of human collaborative activities, we will assume shared intentionality among the agents in the mechanism proposed \cite{tomasello_understanding_2005}.
  
  In Multiplayer sports and games, the team's observation and communication skills can determine a win or lose \cite{stone_teaching_2013}. In the case of a soccer team, spoken language helps to communicate information and intention. Players on one side of a soccer field might not see an opportunity to strike a goal, while another player on the opposite side can. Communicating the observed opportunity could help the team change strategy and increase the probability of striking a goal immediately or in the future. In nature, bees communicate to survive by dancing choreographed patterns, instructing each other to defend the colony, search for food or deploy workers to build and repair the hive \cite{bonabeau_swarm_1999}. Just like a change of colour can indicate i.e. level of aggression in nature, agents can communicate signals of behaviour or intention, or directly send observed information of the environment \cite{van_zoelen_learning_2020}. Collaboration is crucial to succeeding as a group, but the importance of the communicated information might vary and consequently be ignored \cite{srivastava_dropout_2014}. Core questions we ask: Can agents in a MA system learn the importance of communicating information? Subsequently: Can communicated information be used to improve collective performance? Understanding what information is helpful and how to use it to improve collective performance will be the challenge of the experiments conducted in this paper.
  
  \subsection{Contribution}
  In this work, we study communication in the context of a distributed renewable energy grid, an Offshore Wind Farm. Existing offshore wind farms consist of up to 175 wind turbines, covering 90 square kilometres \cite{haddon_wayback_2011}; therefore, wind conditions can vary vastly. Exchanging wind direction information with neighbouring wind turbines can help predict wind change and consequently achieve higher efficiency. Our test bed environment is a fixed size, continuously generated wind field with a random main wind direction. The MA setup consists of eight agents, each controlling one distributed wind turbine, taking actions simultaneously and independently. The collective goal is to maximise efficiency, measured by the cumulative energy generated across all turbines. Orienting a wind turbine against the wind direction generates energy. The environment is partially observable, where each wind turbine agent can only observe its own orientation and wind direction. We can enable a communication layer, which extends each agent's observation space to exchange information with n nearest neighbours \cite{lin_distributed_2016}. Two information exchange modes are available: "broadcasting" and "by Choice". An information signal consists of the local position and wind direction of the sender \cite{sukhbaatar_learning_2016}. Knowing wind direction in the surrounding neighbourhood helps each wind turbine predict future wind change, generating higher cumulative energy over time.
  
  We present a neighbourhood communication system based on a Graph Neural Network (GNN) \cite{scarselli_graph_2009}
  using a message passing framework \cite{gilmer_neural_2017}
  coupled with a Multi-Agent Deep Reinforcement Learning \cite{zhang_multi-agent_2021}
  control mechanism, building on top of a Proximal Policy Optimization (PPO) algorithm
  \cite{schulman_proximal_2017}.
  We demonstrate how training time, performance and scalability of a Single- and no communication Multi-Agent baseline can be surpassed by leveraging communication in a 1. message broadcasting and 2. sending messages by choice setup.

\section{Related Work}
  Crucial ideas and milestones in RL and MA Learning are the basis for MARL \cite{wang_origin_2017}.
  A diversity of fields in research but also industry \cite{leitao_industrial_2015}
  such as game theory, distributed systems, and general Artificial Intelligence are interested in MA systems. Games such as GO \cite{silver_mastering_2016,silver_mastering_2017},
  Chess, Poker, but also autonomous driving \cite{shalev-shwartz_safe_2016}
  and robotics \cite{kober_reinforcement_2013, sukthankar_cooperative_2017}
  are popular testbed environments for RL systems. Most of the mentioned environments include multiple agents in competitive or collaborative settings. Therefore MARL systems can be categorised as such or a mix of both \cite{zhang_multi-agent_2021}.
  Cooperative problem solving by teams composed of people and or computers \cite{jaderberg_human-level_2019}
  requires collaboration and communication. Both are important sub-fields of research in MARL
  \cite{cohen_team_1997,decker_distributed_1987,pynadath_communicative_2002}
  and have a rich historical variety of literature \cite{shoham_multiagent_2009}.
  There have been multiple works on collaboration without communication \cite{matignon_independent_2012, panait_cooperative_2005},
  i.e. by utilising gradient-based distributed policy search methods \cite{peshkin_learning_2000},
  using the same reward function for all agents \cite{lauer_algorithm_2000},
  sharing memory \cite{lowe_multi-agent_2017, pesce_improving_2020, hernandez-leal_survey_2019},
  Parameter Sharing (PS) \cite{sukthankar_cooperative_2017, hernandez-leal_survey_2019},
  Learning with an External Critic (LEC) \cite{whitehead_department_1991},
  or by action replication based on observation \cite{macke_expected_2021}, also called Learning By Watching (LBW) \cite{whitehead_department_1991}.
  
  While we think establishing the notion of collaboration from a non-communicative perspective is important, communication is the focus of the work presented in this paper. Naturally, communication requires a protocol as well as a mode of information transfer. The medium could be i.e. nonverbal communication \cite{mordatch_emergence_2018},
  gesture, colour code, low-level data in the form of binary \cite{berna-koes_communication_2004},
  discrete or continuous signals, or a text-based, structured communication system, also called language. A communication protocol is a set of rules expressed by algorithms and data structures. Communication protocols can be used to share observations in the form of sensory data \cite{mataric_using_1998} or strategies and planned intentions in the form of gradients \cite{foerster_learning_2016}. Giving agents the freedom to develop and learn their own communication protocol can help solve multi-agent problems and discover elegant communication protocols, stated by Foerster et al. (2016) \cite{foerster_learning_2016}.
  For communication to become a choice, it must be part of the action space \cite{xuan_communication_2001}.
  Subsequently, communication needs to come at a cost; otherwise, why would the agent choose not to communicate and prevent itself and others from i.e. a richer observation of the environment. In combination with a collective reward function, individual communication holds collective value. This forces agents to carefully balance between observation and sharing information via communication\cite{macke_expected_2021, pynadath_communicative_2002, zhang_coordinating_2013}.
  Instead of streaming information, there could be real-world limitations that influence research on communication cost. Resources like bandwidth, a fixed count of broadcasting channels, power used, and the width of the network might constrain communication frequency and information data package size or maybe the number of agents able to simultaneously communicate is restricted \cite{kim_learning_2019}.
  There is a social aspect to communication as an individual choice that addresses questions of timing \cite{mirsky_penny_2020, kim_learning_2019}, importance \cite{srivastava_dropout_2014, kim_message-dropout_2019}, and i.e. agent position-dependent selective communication with nearest neighbours only \cite{lin_distributed_2016}.
  Finally, we want to point to work that assumes constant information broadcasting between all or selected agents. It has been argued that a constant broadcasting can provide the learner with shorter latency feedback and auxiliary sources of the experience \cite{whitehead_department_1991}.
  Foerster et al. proposed Reinforced Inter-Agent Learning (RIAL) and Differentiable Inter-Agent Learning (DIAL), which use a discrete communication channel \cite{foerster_learning_2016}. CommNet, in contrast, allows for multiple communication cycles at each timestep, which allows for new or existing agents to join or leave the environment at runtime \cite{sukhbaatar_learning_2016}.
  We believe our work could be classified as a decentralised, partially observable Markov decision process (Dec-POMDP) \cite{oliehoek_decentralized_2012} combined with a message passing \cite{gilmer_neural_2017} communication channel based on a GNN \cite{scarselli_graph_2009}.
  
  Our strategy for MA collaboration through communication builds on previous work including communication at a cost: 1. By choice, as part of the agents' action space \cite{foerster_learning_2016}, and 2. Communication streaming \cite{sukhbaatar_learning_2016}.
  This is coupled with real-world bandwidth constraints, implemented as a nearest neighbour count limitation \cite{lin_distributed_2016}.
  Prior work has investigated RL combined with GNNs \cite{almasan_deep_2020}, but instead of the GNN being the main feedback for the learning agent, our implementation serves as an additional communication layer, not interfering with the environment, rewards or actions. In contrast to Tolstaya et al. (2019) \cite{tolstaya_learning_2021}, our GNN is trained offline and not modified by the RL training process of the learning agent. Pooled neighbourhood messages of ad hoc graphs \cite{tolstaya_learning_2020}, dependent on location and nearest neighbour count, predict future wind directions, while the agent learns when to communicate and pay attention to or ignore messages.

\section{Background}
  \subsection{Proximal Policy Optimization (PPO)}
  The RL algorithm used to train all agents in this work is PPO. The concept of PPO is to estimate a trust region to safely take reasonable learning steps in the right direction, performing gradient ascent. One main concept of the PPO algorithm is Advantage, an estimate for how good an action is compared to the average action for a specific state - a concept used in many Deep RL algorithms including A3C, PPO, and many others \cite{udacity-deeprl_introduction_2019}.
  The Advantage function can be described as follows: $A(s,a) = Q(s,a) - V(s)$, where $s$ is the state and $a$ the action.
  The Q Value (Q Function) is usually denoted as $Q(s,a)$
  and is a measure of the overall expected reward assuming the agent is in state $s$ and performs action $a$, and then continues playing until the end of the episode following some policy $\pi$. Its name is an abbreviation of the word Quality, and defined mathematically as: $\mathcal{Q}(s,a) = \mathbb{E}\left[ \sum_{n=0}^{N} \gamma^n r_n \right]$.
  The State Value Function, usually denoted as $V(s)$ (sometimes with a $\pi$ subscript), is a measure of the overall expected reward assuming the agent is in state $s$ and then continues playing until the end of the episode following some policy $\pi$. It is defined mathematically as: $\mathcal{V}(s) = \mathbb{E}\left[ \sum_{n=0}^{N} \gamma^n r_n \right]$.
  While it does seem similar to the definition of the Q Value, there is an implicit difference: For $n=0$, the reward $r^0$ of $V(s)$ is the expected reward from just being in state $s$, before any action was played, while in the Q Value, $r^0$ is the expected reward after a certain action was played. This difference also yields the Advantage function \cite{zychlinski_complete_2019}.
  The next step is to maintain two separate policy networks: 1. The current policy to be refined $\pi_\theta(a_t|s_t)$, and 2. an older previously established policy, after collecting some experience samples $\pi_{\theta_k}(a_t|s_t)$ and defining the ratio $r_t(\theta) = \frac{\pi_\theta(a_t|s_t)}{\pi_{\theta_k}(a_t|s_t)} = \frac{current\ policy}{old \  policy}$.
  The following is the simplified objective function for performing gradient ascent. It is built to limit how much the current policy deviates from the old policy. If the deviation is too high, the function clips the estimated advantage \cite{achiam_simplified_2018}:
  \begin{quote}
  $\mathcal{L}_{\theta_k}^{CLIP}(\theta) = \underset{s,a\sim\theta_k}{\mathbb{E}} \left[\min{\left( r_t(\theta)A^{\theta_k}(s,a), g(\epsilon,A^{\theta_k}(s,a))\right)}\right]$,
  \newline
  \newline
  where
  \newline
  \newline
  $g(\epsilon,A) = 
  \begin{cases}
    (1 + \epsilon)A,& \text{if } A\geq 0\\
    (1 - \epsilon)A,& \text{otherwise}
  \end{cases}
  $
  \end{quote}
  If the probability ratio between the new policy and the old policy falls outside the range ($1+\epsilon$) and ($1-\epsilon$), the advantage function will be clipped to the value it would have at ($1-\epsilon$) or ($1+\epsilon$). Advantage will not exceed the clipped value. $\epsilon$ was set to 0.2 in the original PPO paper \cite{schulman_proximal_2017}.
  The old policy $\theta_{old}$ will be overridden by the policy that yields the highest sum over all advantage estimates $A_t$ in range of max time step $T$ of a trajectory $\tau \in \mathbb{D}_k$ \cite{openai_proximal_2021}:
  \begin{quote}
  $\theta_{k+1} = arg \underset{\theta_k}{max}\frac{1}{|\mathbb{D}_k|T}\sum_{\tau \in \mathbb{D}_k}\sum_{t=0}^{T}\min\left( \frac{\pi_\theta(a_t|s_t)}{\pi_{\theta_k}(a_t|s_t)},g(\epsilon,A^{\theta_k}(s,a)) \right)$
  \end{quote}
  
  \subsection{Graph Neural Network}
  While GNNs have been introduced by Scarselli F. et al. \cite{scarselli_graph_2009},
  many flavours exist \cite{li_gated_2017, velickovic_graph_2018, defferrard_convolutional_2017}.
  In computer science, a graph is a data structure, and these essential attributes hold for most of them: A graph consists of objects (nodes or vertices) and edges, which are the relationships between two objects. Edges can be directed, i.e. only from node A to B, or undirected, from node A to B and vis versa. Both nodes and edges can hold features.
  
  \begin{figure}[!ht]
    
    \begin{subfigure}{0.5\textwidth}
      \centering
      \includegraphics[width=0.8\linewidth]{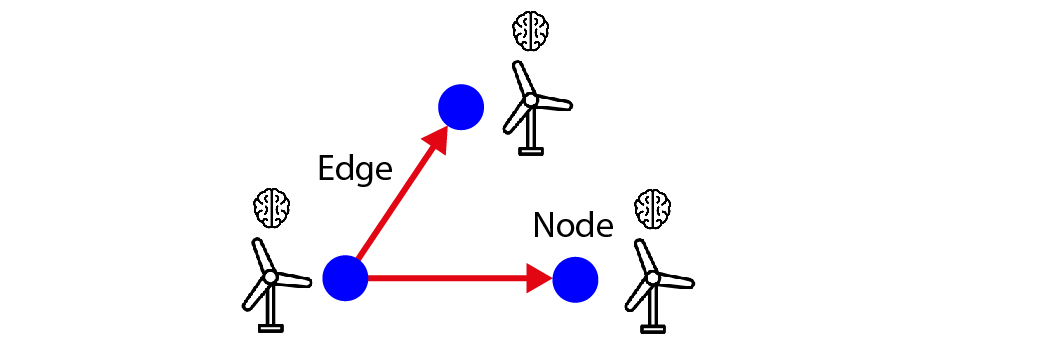}
      \caption{Directed Graph}
      \label{fig:directed-graph}
    \end{subfigure}
    \begin{subfigure}{0.5\textwidth}
      \centering
      \includegraphics[width=0.8\linewidth]{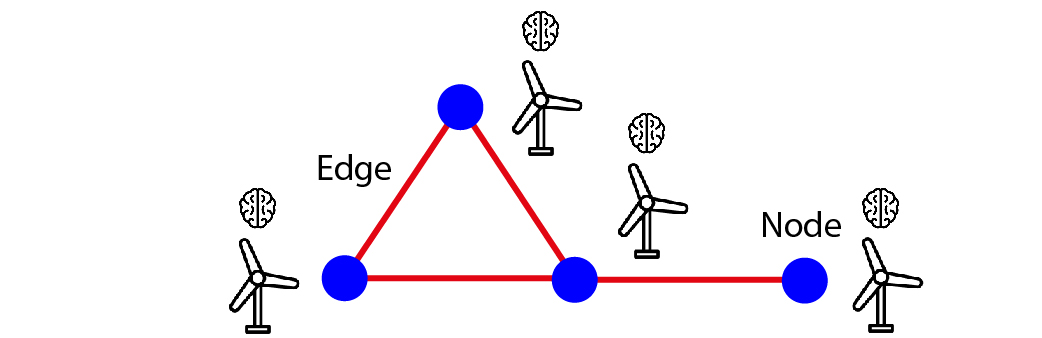}
      \caption{Undirected Graph}
      \label{fig:undirected-graph}
    \end{subfigure}
    \caption{Graph $\mathcal{G}$ consisting of vertices $\mathcal{V}$ (blue dots) and edges $\mathcal{E}$ (red lines): $\mathcal{G} = (\mathcal{V},\mathcal{E})$}
  \end{figure}
  
  A GNN can be used to perform node, edge or graph classification. For example, the state of an object can be predicted using its edges or the existence of an edge using the states of nodes and existing edges. Let us assume we are looking at a wind farm consisting of three wind turbines (Figure \ref{fig:directed-graph}); furthermore, we know the local wind direction of two wind turbines, and we defined that a neighbourhood consists of two nearest neighbours, which we know for all wind turbines. The neighbourhood of turbine 1 is 2 and 3, turbine 2: 1 and 3 and turbine 3: 1 and 2. The local wind direction and the neighbourhood information represent a node state as an n-dimensional vector. This representation can now be learned and used to predict, in our example, the missing local wind direction (Figure \ref{fig:undirected-graph}). The state can be denoted as $h_v$ and consists of features of the edges connecting with $v$, denoted as $x_{co[v]}$. In our example, the embedding (mapped transition of the node) could be the euclidean distance of the neighbouring nodes to $v$, denoted as $h_{ne[v]}$, as well as the features of the neighbouring nodes, denoted as $x_{ne[v]}$. Finally, the function $f$ is the transition function to embed each node onto a n-dimensional space \cite{zhou_graph_2020}:
  \begin{quote}
  $h_v = f(x_v, x_{co[v]}, h_{ne[v], x_{ne[v]}})$
  \end{quote}
  There is a variety of algorithms to define neighbourhood conditions as well as the exchange of information, known as message passing \cite{gilmer_neural_2017}. Popular algorithms to define neighbourhoods are Breadth-First Search (BFS) \cite{burkhardt_optimal_2021}, Depth-First Search (DFS) \cite{kaur_analysis_2012} and random walk based DeepWalk \cite{perozzi_deepwalk_2014}. We used Euclidean distances to find $n$ nearest neighbours.
  The output of the GNN is computed by passing the state $h_v$ and feature $x_v$ to an output function g:
  \begin{quote}
  $o_v = g(h_v, x_v)$
  \end{quote}
  Finally a straight forward gradient descent can be used to formulate loss using the ground truth $t_v$ as well as the output $o_v$ of node $v$:
  \begin{quote}
  $loss = \sum_{i=1}^{p}(t_i - o_i)$
  \end{quote}
  
  \subsection{Multi-Agent Communication}
  
  \begin{figure}[!h]
  \vspace{-0.5cm}
    \includegraphics[width=\linewidth]{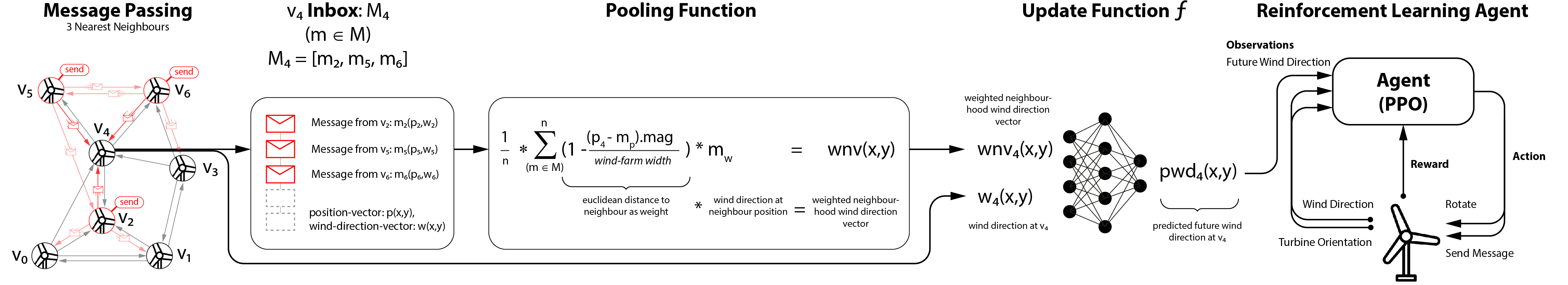} 
    \caption{GNN Message Passing Diagram: Message Passing across neighbourhood; Inbox of the individual wind turbine; Inbox Pooling Function; Neural Network Update Function; GNN future wind prediction as part of the RL Agent observation space. A zoomed-in version of this diagram can be found in the appendix: Figure \ref{fig:GNN_MessagePassing-large}.}
    \label{fig:GNN_MessagePassing}
  \end{figure}
  
  In this work, sending, as part of the action space, and receiving define communication. \textbf{Sending:} In the experiment "Broadcasting" and "by Choice", each wind turbine agent has a neighbourhood of i.e. 3 neighbours. Each agent node $v$ can send a message $m[p(x,y), d(x,y)]$ consisting of its state $h_v$: Normalised local position $p(x,y)$ and wind direction $d(x,y)$, described as a key-value pair, to its neighbourhood $h_{ne[v]}$. Multiple turbines can share the same turbine as a neighbour. \textbf{Receiving:} Each turbine has an infinite large inbox $i[m_1, m_2, m_3, ... , m_n]$, in which messages from other turbines accumulate. The GNN has been trained offline with random wind conditions, predicting future wind direction with its own state and pooled neighbouring wind directions as input, also described in Figure \ref{fig:GNN_MessagePassing}. All received messages are pooled by calculating the mean of the sum over all normalised Euclidean distances between the senders and receiver, multiplied by the senders' local wind direction vector. The weighted neighbourhood wind direction vector at turbine $i$ can be calculated as follows: $wnv(i) = \frac{1}{n}\sum_{m \in M}^{n}(1 - \frac{(p_i - m_p).mag}{wind-farm width})m_w$. $wnv(i)$ and the wind direction at turbine $i$ are the inputs to the neural network of the GNN. Ground truth for learning is the delayed future wind direction at turbine $i$. If there is no message in the inbox, $wnv(i)$ is a zero vector. The trained GNN is then utilised by each wind turbine agent, with the future predicted wind direction as an additional observation. Each agent has to learn how important the future wind prediction is, considering its own state. Sending of a message $m$ happens at time step $t$ and reading at time step $t+1$. All messages are deleted after reading. Receiving is not an action and is happening at every time step $t \in T$. The agent has to learn the benefit of communication, the future wind prediction of messages received, when to send a message to others and how to interact with the environment accordingly to achieve higher collective efficiency.

\section{Methodology}

  \subsection{Wind Farm Environment}
  
  \begin{figure}
    \includegraphics[width=\linewidth]{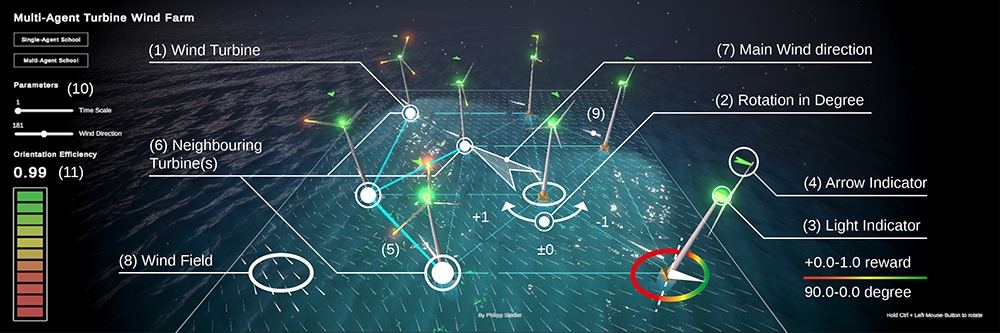}
    \caption{3D Wind Farm Environment: (1) Wind Turbine, (2) Rotation in Degree, (3) Light Indicator, (4) Arrow Indicator, (5) Communication Graph Edge, (6) Neighbouring Turbine(s), (7) Main Wind Direction, (8) Wind Field, (9) Local Wind Direction, (10) Parameter UI: Time Scale, Wind Direction, (11) Performance Display Bar and Number.}
    \label{fig:environment}
  \end{figure}

  We now describe the Offshore Wind Farm environment used for testing and evaluating the Single- and Multi-Agent setups. Figure: \ref{fig:environment} shows the 3D environment including eight distributed wind turbines (1). Each wind turbine can rotate in either of two directions or stand still (2). The wind turbine collective is part of the continuous dynamic environment. A light (3) and an arrow (4) above the turbine indicate its performance. If the turbine orientation is between 0 and 90 degrees against the wind direction, a green to yellow colour is shown and energy is generated. Above 90 degrees against wind direction is indicated with a red colour - the wind will not affect the turbine, and no energy is generated. A cyan coloured line (5) indicates the edges of the communication graph, connecting each wind turbine to their neighbourhood (6). The non-stationary main wind direction (7), rendered in the centre of the field as a large compass needle, is smoothly rotating at random with a speed of -+2 degrees around the z-axis. It is important to note that the wind can change quicker than the turbine can rotate(-+1). As the wind direction changes from the current state to a random next state, the environment is classifiable as stochastic and sequential.
  Based on the main wind direction, the secondary wind field consists of a 2D Perlin Noise field \cite{perlin_image_1985},
  visualised using small white needles distributed on a grid (8). An additional medium-sized white needle (9) at the bottom of each wind turbine shows the local wind direction. A gradient bar and a number ranging from 0.0 to 1.0 show the collective efficiency of the wind farm (11). The turbine orientation and wind direction vector represent a state. Additionally, if the setup can communicate, the future wind prediction by the GNN communication layer is part of a state. Each wind turbine only receives messages from a subset of all turbines, which makes the environment partially observable. The agent reward function per time step is as follows: $t \in T$: ${efficiency} = \sum_{i=0}^{agents}\frac{angle_i(d_i, o_i)}{180} \frac{1.0}{agent count}$, where $o$ is the orientation of the wind turbine, and $d$ the wind direction, and where $angle =
  \begin{cases}
    angle,& \text{if } angle > 0.5\\
    -1.0,& \text{otherwise}
  \end{cases}
  $.
  
  \subsection{Experiments}
  \begin{figure}[H]
  \vspace{-0.5cm}
    \begin{subfigure}{0.24\textwidth}
      \centering
      \includegraphics[width=\linewidth]{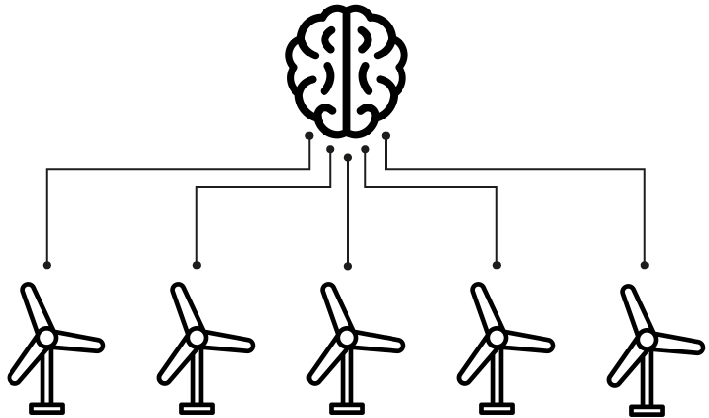}
      \caption{1. Single-Agent}
      \label{fig:SA-setup}
    \end{subfigure}
    \begin{subfigure}{0.24\textwidth}
      \centering
      \includegraphics[width=\linewidth]{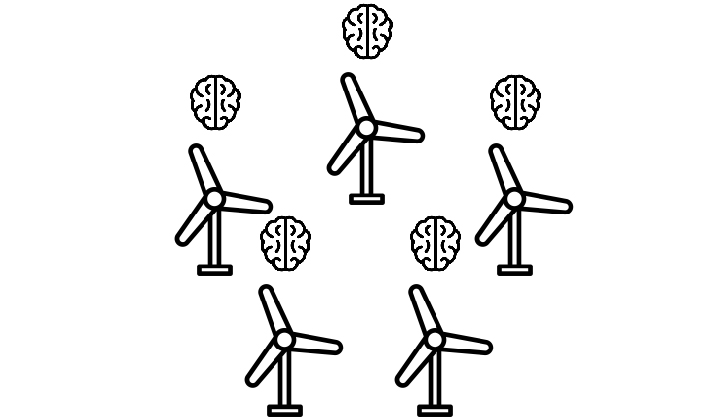}
      \caption{2. Multi-Agent}
      \label{fig:MA-setup}
    \end{subfigure}
    \begin{subfigure}{0.24\textwidth}
      \centering
      \includegraphics[width=\linewidth]{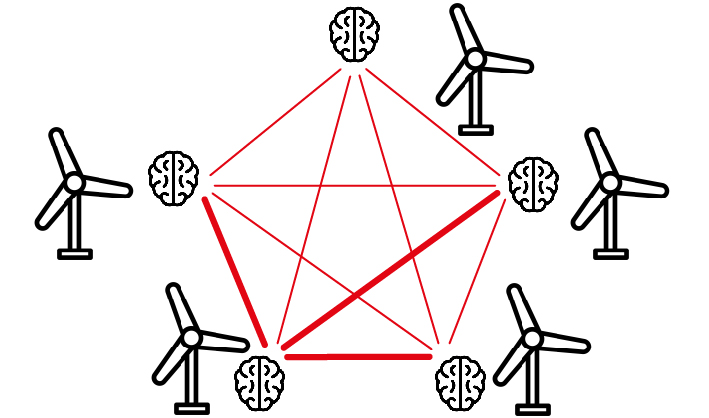}
      \caption{3. Multi-Agent\\ Broadcasting}
      \label{fig:MA-Broadcasting-setup}
    \end{subfigure}
    \begin{subfigure}{0.24\textwidth}
      \centering
      \includegraphics[width=\linewidth]{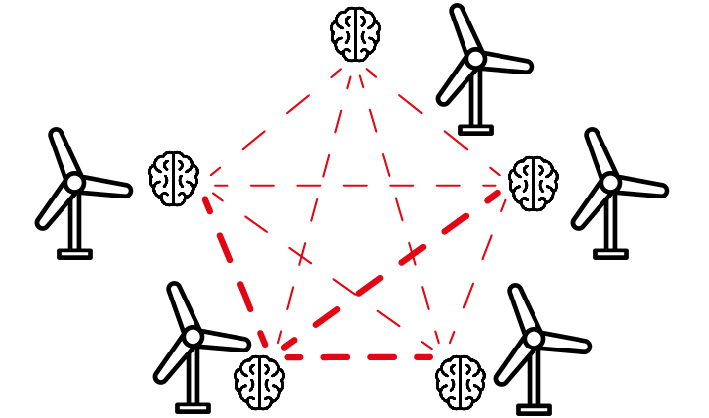}
      \caption{4. Multi-Agent\\ by Choice}
      \label{fig:MA-By-Choice-setup}
    \end{subfigure}
    \caption{Setup Overview Diagram}
  \end{figure}
  
  We have conducted a series of experiments to demonstrate how communication can help a MA setup achieve higher performance with less training time, even when scaling. The following are our four agent setups: First, a centralised SA setup, where one agent controls all turbines (Figure \ref{fig:SA-setup}). The SA setup is one of two baselines to compare and highlight the benefits of communication. In this setup, we are looking at a wind farm consisting of 8, 16 or 24 turbines, which one agent controls. The agent knows the local wind direction as well as the orientation of all turbines. Each wind turbine contributes to a cumulative reward. The max reward the agent can receive per wind turbine per time step is $1.0/n$, where $n$ is the wind turbine count. In an episode of 2000 max time steps, the max cumulative reward is 2000, and the min is -2000 for not generating any energy per episode. Consequently, the second baseline is a decentralised MA setup, where individual agents control all turbines without communication (Figure \ref{fig:MA-setup}). It is converging quicker and achieving higher cumulative rewards in comparison to the SA setup (Figure \ref{fig:training-cumulative-reward}). Following is a decentralised MA setup, with the ability to communicate, automatically broadcasting information at a cost to four nearest neighbours (Figure \ref{fig:MA-Broadcasting-setup}). Each agent's broadcasting cost is a -0.0125 reward per time step, totalling a -25 negative reward per episode. Finally, a decentralised MA setup, also with the ability to communicate at a cost, but by choice (Figure \ref{fig:MA-By-Choice-setup}). The cost is -0.0125 per sending action. So if the agent deems information about a certain circumstance spread worthy, it will send local position and wind direction to four nearest neighbours. Adding cost to communicate forces the agent to learn what circumstances are of interest to the agents' neighbourhood, leading to higher cumulative reward as a collective (Table \ref{performance-table}). Each setup has been trained ten times to capture mean cumulative rewards, training time and time to convergence, for 2000 time steps per episode and 2e6 total time steps. Furthermore, the trained agents have been tested in inference mode on 8, 16 and 24 unseen wind farm settings to measure cumulative reward and  scalability over 2e6 total time steps.
  
  \begin{table}[H]
  \vspace{-0.5cm}
  \caption{Experiment setups and parameters for wind farms including 8 turbines.}
  \label{setup-table}
  \centering
  \begin{tabularx}{\textwidth}{p{2.6cm} p{1cm} p{1cm} p{1.5cm} p{1.1cm} p{1cm} p{1cm} p{1.4cm}}
    \toprule
    \multicolumn{4}{c}{} & \multicolumn{3}{c}{Observation(s)} \\
    \cmidrule(r){5-7}
    Setup & Agent Count & Turbine Count & Neighbour Count & At each Turbine & Total & Stack Size & Action(s) per Agent \\
    \midrule
    Single-Agent     & 1 & 8 & 0 & 4 & 32 & 2 & 24 \\
    Multi-Agent (MA) & 8 & 8 & 0 & 4 & 4  & 2 & 3 \\
    MA Broadcasting  & 8 & 8 & 4 & 6 & 6  & 2 & 3 \\
    MA by Choice     & 8 & 8 & 4 & 6 & 6  & 2 & 5 \\
    \bottomrule
  \end{tabularx}
  \end{table}
  
  \begin{table}[H]
  \vspace{-1.0cm}
  \caption{Experiment results while training and inference mode, including training time, convergence time while training and cumulative rewards for 8, 16 and 24 turbine wind farm setups.}
  \label{performance-table}
  \centering
  \begin{tabularx}{\textwidth}{p{2.6cm} p{1.5cm} p{1.6cm} p{1.9cm} p{1.9cm} p{1.9cm}}
    \toprule
    
    \multicolumn{1}{c}{} & \multicolumn{2}{c}{Training Time (seconds)} & \multicolumn{3}{c}{Inference: Cumulative Reward} \\
    \cmidrule(r){2-3}
    \cmidrule(r){4-6}
    Setup & 2e6 step(s) & Convergence & 8 Turbines & 16 Turbines & 24 Turbines \\
    \multicolumn{1}{c}{} & \multicolumn{1}{l}{(↓ better)} & \multicolumn{1}{l}{(↓ better)} & \multicolumn{1}{l}{(↑ better)} & \multicolumn{1}{l}{(↑ better)} & \multicolumn{1}{l}{(↑ better)} \\
    
    \midrule
    Single-Agent     & 4897.6±21.3           & 4067.3±19.1          & 1596.23±88.18         & 1660.20±77.02         & 1634.99±73.15 \\
    Multi-Agent (MA) & \textbf{1531.6±12.2}  & 1008.5±62.8          & 1733.58±51.06         & 1732.07±40.00         & 1694.74±32.22 \\
    MA Broadcasting  & 1533.6±6.6            & 1044.0±47.5          & 1756.58±54.35         & 1753.38±29.35         & 1734.08±30.61 \\
    MA by Choice     & 1681.6±2.3            & \textbf{1004.0±2.6}  & \textbf{1790.78±56.58}& \textbf{1786.53±20.93}& \textbf{1772.69±18.46} \\
    \bottomrule
  \end{tabularx}
  \end{table}
  
  \subsection{Results}
\begin{figure}[H]
    \centering
    \begin{subfigure}{0.25\textwidth}
      \centering
      \includegraphics[width=\linewidth]{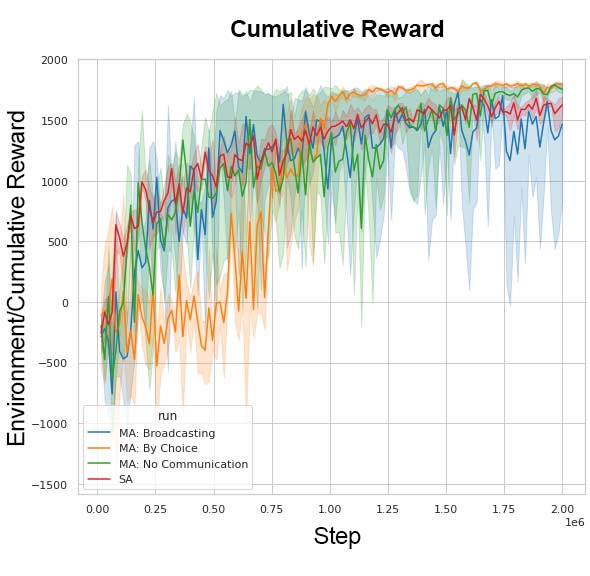}
      \caption{Cumulative Reward}
      \label{fig:training-cumulative-reward}
    \end{subfigure}
    \begin{subfigure}{0.25\textwidth}
      \centering
      \includegraphics[width=\linewidth]{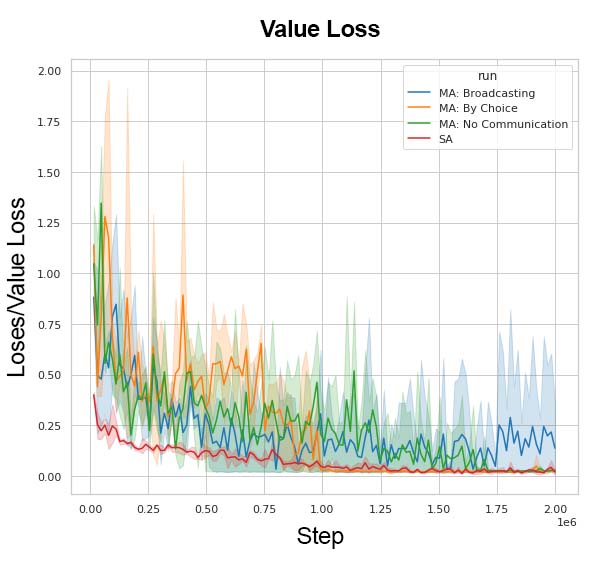}
      \caption{Value Loss}
      \label{fig:training-value-loss}
    \end{subfigure}
    \begin{subfigure}{0.3\textwidth}
      \centering
      \includegraphics[width=\linewidth]{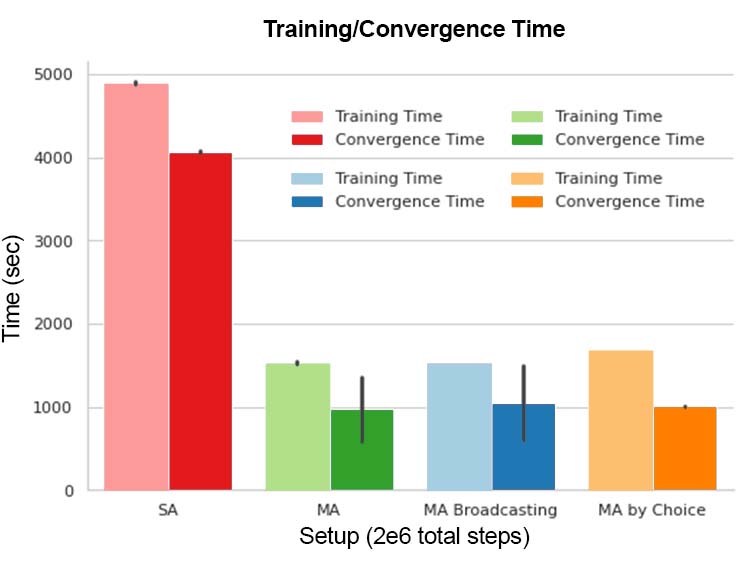}
      \caption{Training/Convergence Time}
      \label{fig:training-time-convergence}
    \end{subfigure}
    \caption{Training: Cumulative reward and value loss per time step. Training and convergence time for 2e6 total time steps. A zoomed-in version of these diagrams can be found in the appendix: Figure \ref{fig:cum-reward-value-loss-large} and \ref{fig:time-scalability-large}.}
  \end{figure}

\begin{figure}[H]
    \centering
    \vspace{-0.8cm}
    \begin{subfigure}{0.25\textwidth}
      \centering
      \includegraphics[width=\linewidth]{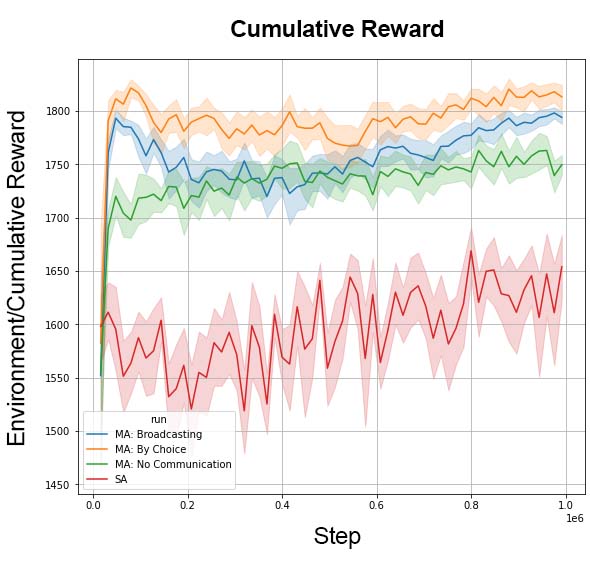}
      \caption{Cumulative Reward}
      \label{fig:cum-reward-inference}
    \end{subfigure}
    \begin{subfigure}{0.25\textwidth}
      \centering
      \includegraphics[width=\linewidth]{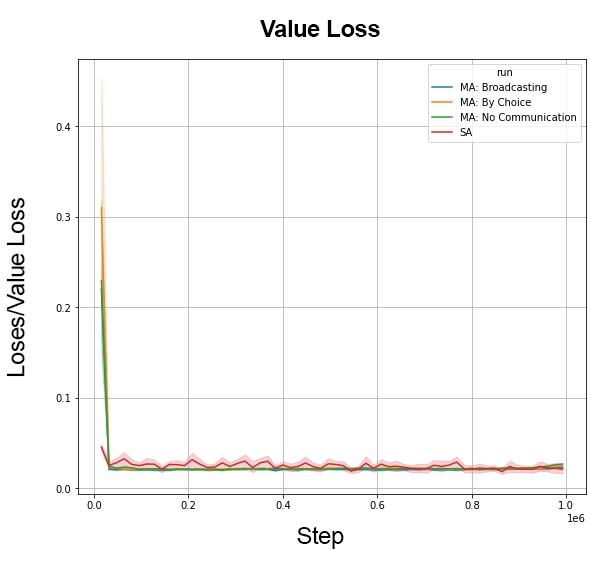}
      \caption{Value Loss}
      \label{fig:value-loss-inference}
    \end{subfigure}
    \begin{subfigure}{0.3\textwidth}
      \centering
      \includegraphics[width=\linewidth]{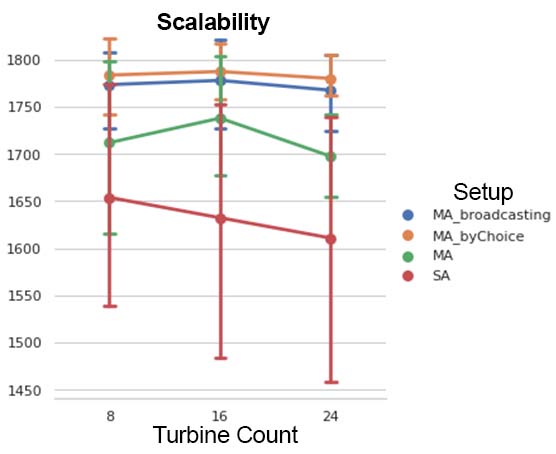}
      \caption{Scalability}
      \label{fig:scalability}
    \end{subfigure}
    \caption{Inference: Default: Cumulative Reward and Value Loss, Random: Scalability. A zoomed-in version of these diagrams can be found in the appendix: Figure \ref{fig:cum-reward-value-loss-inference-large} and \ref{fig:time-scalability-large}.}
  \end{figure}

\begin{figure}[H]
  \vspace{-0.5cm}
    \begin{subfigure}{0.24\textwidth}
      \centering
      \includegraphics[width=\linewidth]{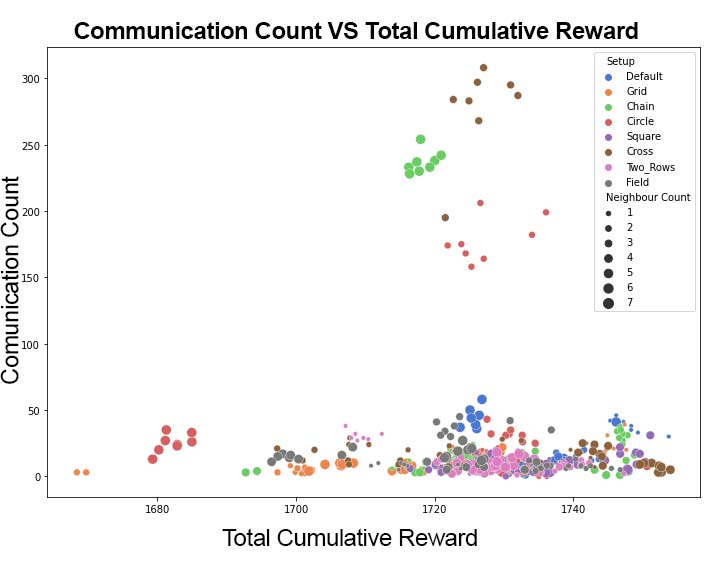}
      \caption{Communication Count VS Total Cumulative \\ Reward}
      \label{fig:com-count-vs-cum-reward}
    \end{subfigure}
    \begin{subfigure}{0.24\textwidth}
      \centering
      \includegraphics[width=\linewidth]{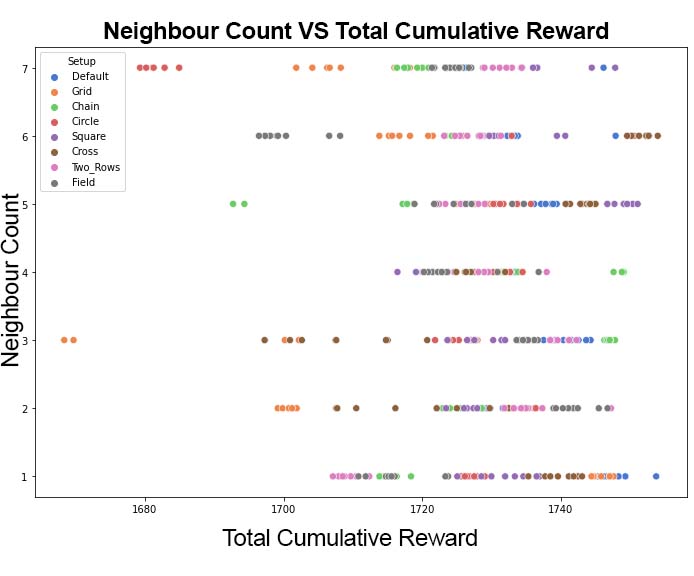}
      \caption{Neighbour Count VS Total Cumulative \\ Reward}
      \label{fig:n-count-vs-cum-reward}
    \end{subfigure}
    \begin{subfigure}{0.24\textwidth}
      \centering
      \includegraphics[width=\linewidth]{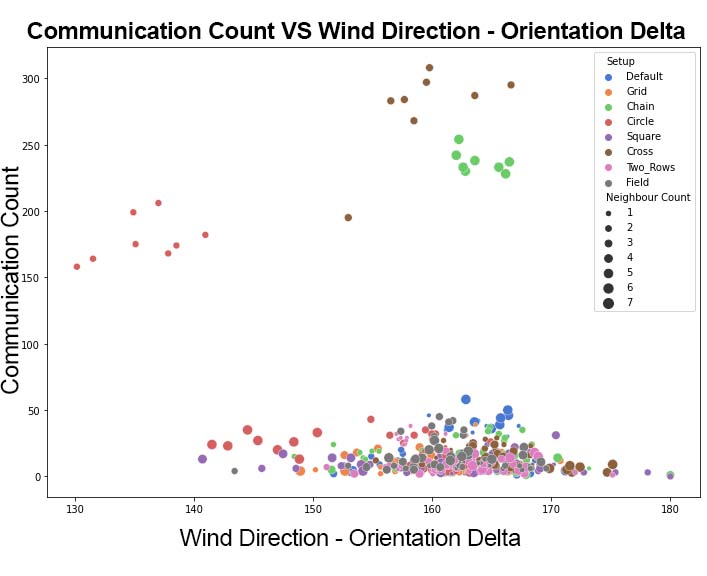}
      \caption{Communication Count VS Wind Direction - \\ Turbine Orientation Delta}
      \label{fig:com-count-vs-wind-dir-orientation}
    \end{subfigure}
    \begin{subfigure}{0.24\textwidth}
      \centering
      \includegraphics[width=\linewidth]{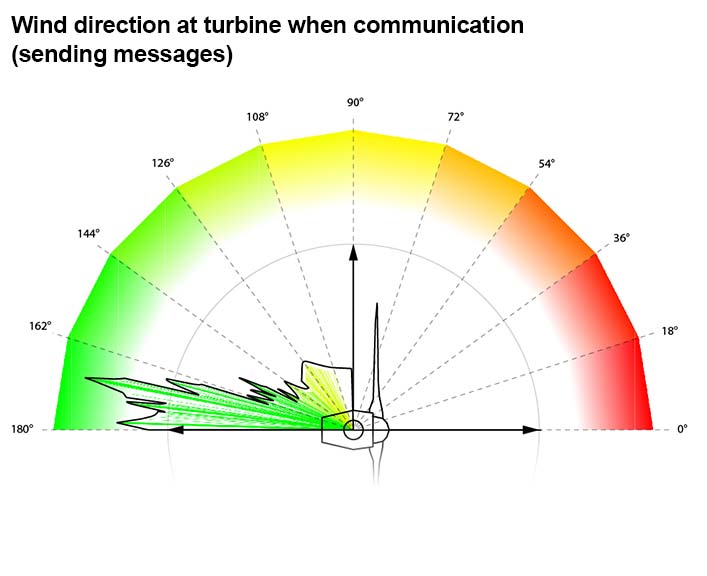}
      \caption{Wind direction when communicating}
      \label{fig:wind-dir-at-turbine-when-com}
    \end{subfigure}
    \caption{These diagrams indicate if communication and neighbour count is related to performance, and what wind direction angle is present, at the moment of communication. A zoomed-in version of these diagrams can be found in the appendix: Figure \ref{fig:com-count-vs-cum-reward-large}, \ref{fig:n-count-cum-reward-large}, \ref{fig:com-count-vs-wind-dir-orientation-large}, \ref{fig:wind-dir-at-turbine-when-com-large}.}
  \end{figure}

  We have demonstrated that an individual agent in a MA setup, given the ability to communicate, reduces total time for training and convergence, increases the collective performance through higher cumulative reward in inference mode and shows high stability while scaling.
  MA by Choice is converging at 1e6 time steps, while MA Broadcasting, at approx. 1.5e6, MA no Communication at approx. 1.6e6 - almost identical but at a lower cumulative reward as SA (Figure \ref{fig:training-cumulative-reward}, Table \ref{performance-table}).
  When testing the trained agent setups in inference mode, the MA setups outperform the SA setup. Despite possible total cost of -25 reward for communication for MA Broadcasting and MA By Choice setups, both surpass the MA no Communication and SA setup. Giving the agent the choice to communicate increases the cumulative reward by roughly 2\% in comparison to Broadcasting (Figure \ref{fig:cum-reward-inference}, Table \ref{performance-table}). This has been tested on 20 inference runs, each with 1e6 time steps total.
  Scaling the environment turbine count from 8 to 16 and 24 turbines per wind farm, shows how much better both MA setups with the ability to communicate perform (Figure \ref{fig:scalability}, Table \ref{performance-table}). Both MA by Choice and Broadcasting agents developed a sense of neighbourhood and an understand the benefit of communication.
  Figure \ref{fig:com-count-vs-cum-reward} gives a clear indication that high communication count is related to high performance. We have investigated what the most beneficial neighbour count is and found that 4 neighbours have the highest total cumulative reward (Figure \ref{fig:n-count-vs-cum-reward}). Furthermore we can say that a wind turbine is communicating more often when performing well (Figure \ref{fig:com-count-vs-wind-dir-orientation} and \ref{fig:wind-dir-at-turbine-when-com-large}).

\section{Discussion and Future Work}

Communication protocols are hard-coded in this work, which the agent could develop. The GNN could be trained online in correlation with the reward function of the RL agent. Further questions we might ask: Could we give agents the ability to develop urgency? Moreover, how do distribution patterns affect the information propagation through the wind farm? In addition, the physics of the environment is simplified, mass and turbulence are not considered, and the wind is in 2 dimensions only. We use nearest neighbours to generate neighbourhoods, but other algorithms build neighbourhood graphs for communication, as mentioned in the GNN section.

\newpage
\begin{ack}
We want to thank Serkan Cabi, Aleksandra Faust, Panagiotis Tigas, Ilinca Barsan, Philipp C. Paulus, Jasmin Arensmeier and Thore Graepel, without their constant support, patience, guidance and encouragement this would not have been possible. We also want to thank the reviewing committee for their efforts and critic, which was very helpful to push the work one step further.
Further information, video material and an interactive web-app can be found at: \url{https://ai.philippsiedler.com/neurips2021-cooperativeai-gnn-marl-wind-farm/}.
\end{ack}

{
\small
\bibliography{references}

\begin{thebibliography}{10}

\bibitem{cohen_team_1997}
Philip Cohen, Hector Levesque, and Ira Smith.
\newblock On {Team} {Formation}.
\newblock In {\em Contemporary {Action} {Theory}. {Synthese}}, pages 87--114.
  Kluwer Academic Publishers, 1997.

\bibitem{guestrin_coordinated_2002}
Carlos Guestrin, Michail Lagoudakis, and Ronald Parr.
\newblock Coordinated {Reinforcement} {Learning}.
\newblock In {\em In {Proceedings} of the {ICML}-2002 {The} {Nineteenth}
  {International} {Conference} on {Machine} {Learning}}, pages 227--234, 2002.

\bibitem{decker_distributed_1987}
Keith~S. Decker.
\newblock Distributed problem-solving techniques: {A} survey.
\newblock {\em IEEE Transactions on Systems, Man, \& Cybernetics},
  17(5):729--740, 1987.
\newblock Place: US Publisher: Institute of Electrical \& Electronics Engineers
  Inc.

\bibitem{panait_cooperative_2005}
Liviu Panait and Sean Luke.
\newblock Cooperative {Multi}-{Agent} {Learning}: {The} {State} of the {Art}.
\newblock {\em Autonomous Agents and Multi-Agent Systems}, 11(3):387--434,
  November 2005.

\bibitem{mataric_using_1998}
MAJA~J. MATARIC.
\newblock Using communication to reduce locality in distributed multiagent
  learning.
\newblock {\em Journal of Experimental \& Theoretical Artificial Intelligence},
  10(3):357--369, July 1998.
\newblock Publisher: Taylor \& Francis \_eprint:
  https://doi.org/10.1080/095281398146806.

\bibitem{ravula_ad_2019}
Manish Ravula, Shani Alkoby, and Peter Stone.
\newblock Ad {Hoc} {Teamwork} {With} {Behavior} {Switching} {Agents}.
\newblock In {\em Proceedings of the {Twenty}-{Eighth} {International} {Joint}
  {Conference} on {Artificial} {Intelligence}}, pages 550--556. International
  Joint Conferences on Artificial Intelligence Organization, July 2019.

\bibitem{hernandez-leal_survey_2019-1}
Pablo Hernandez-Leal, Bilal Kartal, and Matthew~E. Taylor.
\newblock A {Survey} and {Critique} of {Multiagent} {Deep} {Reinforcement}
  {Learning}.
\newblock {\em Autonomous Agents and Multi-Agent Systems}, 33(6):750--797,
  November 2019.
\newblock arXiv: 1810.05587.

\bibitem{tomasello_understanding_2005}
Michael Tomasello, Malinda Carpenter, Josep Call, Tanya Behne, and Henrike
  Moll.
\newblock Understanding and sharing intentions: the origins of cultural
  cognition.
\newblock {\em The Behavioral and Brain Sciences}, 28(5):675--691; discussion
  691--735, October 2005.

\bibitem{stone_teaching_2013}
Peter Stone, Gal~A. Kaminka, Sarit Kraus, Jeffrey~S. Rosenschein, and Noa
  Agmon.
\newblock Teaching and leading an ad hoc teammate: {Collaboration} without
  pre-coordination.
\newblock {\em Artificial Intelligence}, 203:35--65, October 2013.

\bibitem{bonabeau_swarm_1999}
Eric Bonabeau, Marco Dorigo, and Guy Theraulaz.
\newblock {\em Swarm {Intelligence}: {From} {Natural} to {Artificial}
  {Systems}}.
\newblock Santa {Fe} {Institute} {Studies} on the {Sciences} of {Complexity}.
  Oxford University Press, New York, 1999.

\bibitem{van_zoelen_learning_2020}
Emma~M. van Zoelen, Anita Cremers, Frank P.~M. Dignum, Jurriaan van Diggelen,
  and Marieke~M. Peeters.
\newblock Learning to {Communicate} {Proactively} in {Human}-{Agent} {Teaming}.
\newblock In Fernando De~La~Prieta, Philippe Mathieu, Jaime~Andrés
  Rincón~Arango, Alia El~Bolock, Elena Del~Val, Jaume Jordán~Prunera, João
  Carneiro, Rubén Fuentes, Fernando Lopes, and Vicente Julian, editors, {\em
  Highlights in {Practical} {Applications} of {Agents}, {Multi}-{Agent}
  {Systems}, and {Trust}-worthiness. {The} {PAAMS} {Collection}},
  Communications in {Computer} and {Information} {Science}, pages 238--249,
  Cham, 2020. Springer International Publishing.

\bibitem{srivastava_dropout_2014}
Nitish Srivastava, Geoﬀrey Hinton, Alex Krizhevsky, Ilya Sutskever, and
  Ruslan Salakhutdinov.
\newblock Dropout: {A} {Simple} {Way} to {Prevent} {Neural} {Networks} from
  {Overﬁtting}.
\newblock {\em Journal of Machine Learning Research}, 15 (2014) 1929-1958:30,
  2014.

\bibitem{haddon_wayback_2011}
Joanne Haddon.
\newblock Wayback {Machine}, July 2011.

\bibitem{lin_distributed_2016}
Peng Lin, Wei Ren, and Yongduan Song.
\newblock Distributed multi-agent optimization subject to nonidentical
  constraints and communication delays.
\newblock {\em Automatica}, 65:120--131, March 2016.

\bibitem{sukhbaatar_learning_2016}
Sainbayar Sukhbaatar, Arthur Szlam, and Rob Fergus.
\newblock Learning {Multiagent} {Communication} with {Backpropagation}.
\newblock {\em arXiv:1605.07736 [cs]}, October 2016.
\newblock arXiv: 1605.07736.

\bibitem{scarselli_graph_2009}
Franco Scarselli, Marco Gori, Ah~Chung Tsoi, Markus Hagenbuchner, and Gabriele
  Monfardini.
\newblock The graph neural network model.
\newblock {\em IEEE Transactions on Neural Networks}, 2009.

\bibitem{gilmer_neural_2017}
Justin Gilmer, Samuel~S. Schoenholz, Patrick~F. Riley, Oriol Vinyals, and
  George~E. Dahl.
\newblock Neural {Message} {Passing} for {Quantum} {Chemistry}.
\newblock {\em arXiv:1704.01212 [cs]}, June 2017.
\newblock arXiv: 1704.01212.

\bibitem{zhang_multi-agent_2021}
Kaiqing Zhang, Zhuoran Yang, and Tamer Başar.
\newblock Multi-{Agent} {Reinforcement} {Learning}: {A} {Selective} {Overview}
  of {Theories} and {Algorithms}.
\newblock {\em arXiv:1911.10635 [cs, stat]}, April 2021.
\newblock arXiv: 1911.10635.

\bibitem{schulman_proximal_2017}
John Schulman, Filip Wolski, Prafulla Dhariwal, Alec Radford, and Oleg Klimov.
\newblock Proximal {Policy} {Optimization} {Algorithms}.
\newblock {\em arXiv:1707.06347 [cs]}, August 2017.
\newblock arXiv: 1707.06347.

\bibitem{wang_origin_2017}
Haohan Wang and Bhiksha Raj.
\newblock On the {Origin} of {Deep} {Learning}.
\newblock {\em arXiv:1702.07800 [cs, stat]}, March 2017.
\newblock arXiv: 1702.07800.

\bibitem{leitao_industrial_2015}
Paulo Leitão and Stamatis Karnouskos.
\newblock {\em Industrial {Agents}: {Emerging} {Applications} of {Software}
  {Agents} in {Industry}}.
\newblock Elsevier, March 2015.

\bibitem{silver_mastering_2016}
David Silver, Aja Huang, Chris~J. Maddison, Arthur Guez, Laurent Sifre, George
  van~den Driessche, Julian Schrittwieser, Ioannis Antonoglou, Veda
  Panneershelvam, Marc Lanctot, Sander Dieleman, Dominik Grewe, John Nham, Nal
  Kalchbrenner, Ilya Sutskever, Timothy Lillicrap, Madeleine Leach, Koray
  Kavukcuoglu, Thore Graepel, and Demis Hassabis.
\newblock Mastering the game of {Go} with deep neural networks and tree search.
\newblock {\em Nature}, 529(7587):484--489, January 2016.
\newblock Bandiera\_abtest: a Cg\_type: Nature Research Journals Number: 7587
  Primary\_atype: Research Publisher: Nature Publishing Group Subject\_term:
  Computational science;Computer science;Reward Subject\_term\_id:
  computational-science;computer-science;reward.

\bibitem{silver_mastering_2017}
David Silver, Julian Schrittwieser, Karen Simonyan, Ioannis Antonoglou, Aja
  Huang, Arthur Guez, Thomas Hubert, Lucas Baker, Matthew Lai, Adrian Bolton,
  Yutian Chen, Timothy Lillicrap, Fan Hui, Laurent Sifre, George van~den
  Driessche, Thore Graepel, and Demis Hassabis.
\newblock Mastering the game of {Go} without human knowledge.
\newblock {\em Nature}, 550(7676):354--359, October 2017.
\newblock Bandiera\_abtest: a Cg\_type: Nature Research Journals Number: 7676
  Primary\_atype: Research Publisher: Nature Publishing Group Subject\_term:
  Computational science;Computer science;Reward Subject\_term\_id:
  computational-science;computer-science;reward.

\bibitem{shalev-shwartz_safe_2016}
Shai Shalev-Shwartz, Shaked Shammah, and Amnon Shashua.
\newblock Safe, {Multi}-{Agent}, {Reinforcement} {Learning} for {Autonomous}
  {Driving}.
\newblock {\em arXiv:1610.03295 [cs, stat]}, October 2016.
\newblock arXiv: 1610.03295.

\bibitem{kober_reinforcement_2013}
Jens Kober, J~Andrew Bagnell, and Jan Peters.
\newblock Reinforcement {Learning} in {Robotics}: {A} {Survey}.
\newblock {\em The International Journal of Robotics Research
  32(11):1238-1274}, page~38, 2013.

\bibitem{sukthankar_cooperative_2017}
Jayesh~K. Gupta, Maxim Egorov, and Mykel Kochenderfer.
\newblock Cooperative {Multi}-agent {Control} {Using} {Deep} {Reinforcement}
  {Learning}.
\newblock In Gita Sukthankar and Juan~A. Rodriguez-Aguilar, editors, {\em
  Autonomous {Agents} and {Multiagent} {Systems}}, volume 10642, pages 66--83.
  Springer International Publishing, Cham, 2017.
\newblock Series Title: Lecture Notes in Computer Science.

\bibitem{jaderberg_human-level_2019}
Max Jaderberg, Wojciech~M. Czarnecki, Iain Dunning, Luke Marris, Guy Lever,
  Antonio~Garcia Castañeda, Charles Beattie, Neil~C. Rabinowitz, Ari~S.
  Morcos, Avraham Ruderman, Nicolas Sonnerat, Tim Green, Louise Deason, Joel~Z.
  Leibo, David Silver, Demis Hassabis, Koray Kavukcuoglu, and Thore Graepel.
\newblock Human-level performance in {3D} multiplayer games with
  population-based reinforcement learning.
\newblock {\em Science}, 364(6443):859--865, May 2019.
\newblock Publisher: American Association for the Advancement of Science.

\bibitem{pynadath_communicative_2002}
D.~V. Pynadath and M.~Tambe.
\newblock The {Communicative} {Multiagent} {Team} {Decision} {Problem}:
  {Analyzing} {Teamwork} {Theories} and {Models}.
\newblock {\em Journal of Artificial Intelligence Research}, 16:389--423, June
  2002.
\newblock arXiv: 1106.4569.

\bibitem{shoham_multiagent_2009}
Yoav Shoham.
\newblock {\em Multiagent {Systems}: {Algorithmic}, {Game}-{Theoretic}, and
  {Logical} {Foundations}}.
\newblock Cambridge University Press, 2009.

\bibitem{matignon_independent_2012}
Laëtitia Matignon, Guillaume~J. Laurent, and Nadine Le~Fort-Piat.
\newblock Independent reinforcement learners in cooperative {Markov} games: a
  survey regarding coordination problems.
\newblock {\em Knowledge Engineering Review}, 27(1):1--31, March 2012.
\newblock Publisher: Cambridge University Press (CUP).

\bibitem{peshkin_learning_2000}
Leonid Peshkin, Kee-Eung Kim, Nicolas Meuleau, and Leslie Pack~Kaelnling.
\newblock Learning to {Cooperate} via {Policy} {Search}, 2000.

\bibitem{lauer_algorithm_2000}
Martin Lauer and Martin Riedmiller.
\newblock An {Algorithm} for {Distributed} {Reinforcement} {Learning} in
  {Cooperative} {Multi}-{Agent} {Systems}.
\newblock In {\em In {Proceedings} of the {Seventeenth} {International}
  {Conference} on {Machine} {Learning}}, pages 535--542. Morgan Kaufmann, 2000.

\bibitem{lowe_multi-agent_2017}
Ryan Lowe, YI~WU, Aviv Tamar, Jean Harb, OpenAI Pieter~Abbeel, and Igor
  Mordatch.
\newblock Multi-{Agent} {Actor}-{Critic} for {Mixed}
  {Cooperative}-{Competitive} {Environments}.
\newblock In {\em Advances in {Neural} {Information} {Processing} {Systems}},
  volume~30. Curran Associates, Inc., 2017.

\bibitem{pesce_improving_2020}
Emanuele Pesce and Giovanni Montana.
\newblock Improving {Coordination} in {Small}-{Scale} {Multi}-{Agent} {Deep}
  {Reinforcement} {Learning} through {Memory}-driven {Communication}.
\newblock {\em Machine Learning}, 109(9-10):1727--1747, September 2020.
\newblock arXiv: 1901.03887.

\bibitem{hernandez-leal_survey_2019}
Pablo Hernandez-Leal, Michael Kaisers, Tim Baarslag, and Enrique~Munoz de~Cote.
\newblock A {Survey} of {Learning} in {Multiagent} {Environments}: {Dealing}
  with {Non}-{Stationarity}.
\newblock {\em arXiv:1707.09183 [cs]}, March 2019.
\newblock arXiv: 1707.09183.

\bibitem{whitehead_department_1991}
D~Whitehead.
\newblock Department of {Computer} {Science} {University} of {Rochester}
  {Rochester}, {NY} 14627 email: white@cs.rochester.edu.
\newblock In {\em {AAAI}-91 {Proceedings}}, page~7. AAAI (www.aaai.org), 1991.

\bibitem{macke_expected_2021}
William Macke, Reuth Mirsky, and Peter Stone.
\newblock Expected {Value} of {Communication} for {Planning} in {Ad} {Hoc}
  {Teamwork}.
\newblock {\em arXiv:2103.01171 [cs]}, March 2021.
\newblock arXiv: 2103.01171.

\bibitem{mordatch_emergence_2018}
Igor Mordatch and Pieter Abbeel.
\newblock Emergence of {Grounded} {Compositional} {Language} in {Multi}-{Agent}
  {Populations}.
\newblock {\em Proceedings of the AAAI Conference on Artificial Intelligence},
  32(1), April 2018.
\newblock Number: 1.

\bibitem{berna-koes_communication_2004}
M.~Berna-Koes, I.~Nourbakhsh, and K.~Sycara.
\newblock Communication efficiency in multi-agent systems.
\newblock In {\em {IEEE} {International} {Conference} on {Robotics} and
  {Automation}, 2004. {Proceedings}. {ICRA} '04. 2004}, volume~3, pages
  2129--2134 Vol.3, April 2004.
\newblock ISSN: 1050-4729.

\bibitem{foerster_learning_2016}
Jakob Foerster, Ioannis~Alexandros Assael, Nando de~Freitas, and Shimon
  Whiteson.
\newblock Learning to {Communicate} with {Deep} {Multi}-{Agent} {Reinforcement}
  {Learning}.
\newblock In {\em Advances in {Neural} {Information} {Processing} {Systems}},
  volume~29. Curran Associates, Inc., 2016.

\bibitem{xuan_communication_2001}
Ping Xuan, Victor Lesser, and Shlomo Zilberstein.
\newblock Communication decisions in multi-agent cooperation: model and
  experiments.
\newblock In {\em Proceedings of the fifth international conference on
  {Autonomous} agents}, {AGENTS} '01, pages 616--623, New York, NY, USA, May
  2001. Association for Computing Machinery.

\bibitem{zhang_coordinating_2013}
Chongjie Zhang and Victor Lesser.
\newblock Coordinating {Multi}-{Agent} {Reinforcement} {Learning} with
  {Limited} {Communication}.
\newblock In {\em Proceedings of the 12th {International} {Conference} on
  Autonomous {Agents} and {Multiagent} {Systems} ({AAMAS} 2013)}, page~8.
  AAMAS, 2013.

\bibitem{kim_learning_2019}
Daewoo Kim, Sangwoo Moon, David Hostallero, Wan~Ju Kang, Taeyoung Lee,
  Kyunghwan Son, and Yung Yi.
\newblock Learning to {Schedule} {Communication} in {Multi}-agent
  {Reinforcement} {Learning}.
\newblock {\em arXiv:1902.01554 [cs]}, February 2019.
\newblock arXiv: 1902.01554.

\bibitem{mirsky_penny_2020}
Reuth Mirsky, William Macke, Andy Wang, Harel Yedidsion, and Peter Stone.
\newblock A {Penny} for {Your} {Thoughts}: {The} {Value} of {Communication} in
  {Ad} {Hoc} {Teamwork}.
\newblock In {\em Proceedings of the {Twenty}-{Ninth} {International} {Joint}
  {Conference} on {Artificial} {Intelligence} ({IJCAI}-20}, volume~1, pages
  254--260, July 2020.
\newblock ISSN: 1045-0823.

\bibitem{kim_message-dropout_2019}
Woojun Kim, Myungsik Cho, and Youngchul Sung.
\newblock Message-{Dropout}: {An} {Efficient} {Training} {Method} for
  {Multi}-{Agent} {Deep} {Reinforcement} {Learning}.
\newblock {\em Proceedings of the AAAI Conference on Artificial Intelligence},
  33(01):6079--6086, July 2019.
\newblock Number: 01.

\bibitem{oliehoek_decentralized_2012}
Frans~A. Oliehoek.
\newblock Decentralized {POMDPs}.
\newblock In Marco Wiering and Martijn van Otterlo, editors, {\em Reinforcement
  {Learning}: {State}-of-the-{Art}}, Adaptation, {Learning}, and
  {Optimization}, pages 471--503. Springer, Berlin, Heidelberg, 2012.

\bibitem{almasan_deep_2020}
Paul Almasan, José Suárez-Varela, Arnau Badia-Sampera, Krzysztof Rusek, Pere
  Barlet-Ros, and Albert Cabellos-Aparicio.
\newblock Deep {Reinforcement} {Learning} meets {Graph} {Neural} {Networks}:
  exploring a routing optimization use case.
\newblock {\em arXiv:1910.07421 [cs]}, February 2020.
\newblock arXiv: 1910.07421.

\bibitem{tolstaya_learning_2021}
Ekaterina Tolstaya, Landon Butler, Daniel Mox, James Paulos, Vijay Kumar, and
  Alejandro Ribeiro.
\newblock Learning {Connectivity} for {Data} {Distribution} in {Robot} {Teams}.
\newblock {\em arXiv:2103.05091 [cs]}, July 2021.
\newblock arXiv: 2103.05091.

\bibitem{tolstaya_learning_2020}
Ekaterina Tolstaya, Fernando Gama, James Paulos, George Pappas, Vijay Kumar,
  and Alejandro Ribeiro.
\newblock Learning {Decentralized} {Controllers} for {Robot} {Swarms} with
  {Graph} {Neural} {Networks}.
\newblock In {\em Proceedings of the {Conference} on {Robot} {Learning}}, pages
  671--682. PMLR, May 2020.
\newblock ISSN: 2640-3498.

\bibitem{udacity-deeprl_introduction_2019}
{Udacity-DeepRL}.
\newblock An {Introduction} to {Proximal} {Policy} {Optimization} ({PPO}) in
  {Deep} {Reinforcement} {Learning}, April 2019.

\bibitem{zychlinski_complete_2019}
Shaked Zychlinski.
\newblock The {Complete} {Reinforcement} {Learning} {Dictionary}, November
  2019.

\bibitem{achiam_simplified_2018}
Joshua Achiam.
\newblock Simplified {PPO}-{Clip} {Objective}, July 2018.

\bibitem{openai_proximal_2021}
Spinning~Up OpenAI.
\newblock Proximal {Policy} {Optimization} — {Spinning} {Up} documentation,
  2021.

\bibitem{li_gated_2017}
Yujia Li, Daniel Tarlow, Marc Brockschmidt, and Richard Zemel.
\newblock Gated {Graph} {Sequence} {Neural} {Networks}.
\newblock {\em arXiv:1511.05493 [cs, stat]}, September 2017.
\newblock arXiv: 1511.05493.

\bibitem{velickovic_graph_2018}
Petar Veličković, Guillem Cucurull, Arantxa Casanova, Adriana Romero, Pietro
  Liò, and Yoshua Bengio.
\newblock Graph {Attention} {Networks}.
\newblock {\em arXiv:1710.10903 [cs, stat]}, February 2018.
\newblock arXiv: 1710.10903.

\bibitem{defferrard_convolutional_2017}
Michaël Defferrard, Xavier Bresson, and Pierre Vandergheynst.
\newblock Convolutional {Neural} {Networks} on {Graphs} with {Fast} {Localized}
  {Spectral} {Filtering}.
\newblock {\em arXiv:1606.09375 [cs, stat]}, February 2017.
\newblock arXiv: 1606.09375.

\bibitem{zhou_graph_2020}
Jie Zhou, Ganqu Cui, Shengding Hu, Zhengyan Zhang, Cheng Yang, Zhiyuan Liu,
  Lifeng Wang, Changcheng Li, and Maosong Sun.
\newblock Graph neural networks: {A} review of methods and applications.
\newblock {\em AI Open}, 1:57--81, 2020.

\bibitem{burkhardt_optimal_2021}
Paul Burkhardt.
\newblock Optimal algebraic {Breadth}-{First} {Search} for sparse graphs.
\newblock {\em ACM Transactions on Knowledge Discovery from Data}, 15(5):1--19,
  June 2021.
\newblock arXiv: 1906.03113.

\bibitem{kaur_analysis_2012}
N.~Kaur and D.~Garg.
\newblock Analysis of the {Depth} {First} {Search} {Algorithms}.
\newblock {\em undefined}, 2012.

\bibitem{perozzi_deepwalk_2014}
Bryan Perozzi, Rami Al-Rfou, and Steven Skiena.
\newblock {DeepWalk}: {Online} {Learning} of {Social} {Representations}.
\newblock {\em Proceedings of the 20th ACM SIGKDD international conference on
  Knowledge discovery and data mining}, pages 701--710, August 2014.
\newblock arXiv: 1403.6652.

\bibitem{perlin_image_1985}
Ken Perlin.
\newblock An image synthesizer.
\newblock {\em ACM SIGGRAPH Computer Graphics}, 19(3):287--296, July 1985.

\end{thebibliography}
}

\appendix
\section{Hyperparameters}
\subsection{Hyperparameters used in all experiments presented in this paper}
\begin{verbatim}
behaviors:
  TurbineAgent:
    trainer_type: ppo
    hyperparameters:
      batch_size: 32
      buffer_size: 256
      learning_rate: 0.0003
      beta: 0.005
      epsilon: 0.2
      lambd: 0.95
      num_epoch: 3
      learning_rate_schedule: linear
    network_settings:
      normalize: false
      hidden_units: 20
      num_layers: 3
      vis_encode_type: simple
    reward_signals:
      extrinsic:
        gamma: 0.9
        strength: 1.0
    keep_checkpoints: 5
    max_steps: 2000000
    time_horizon: 3
    summary_freq: 16000
    threaded: true
\end{verbatim}

\newpage
\subsection{Hyperparameter Description}

\begin{table}[h]
  \begin{tabular}{p{0.3\textwidth}p{0.3\textwidth}p{0.3\textwidth}}
    \toprule
    Hyperparameter & Typical Range & Description\\
    \midrule
    Gamma & $0.8-0.995$ & discount factor for future rewards\\
    Lambda & $0.9-0.95$ & used when calculating the Generalized Advantage Estimate (GAE)\\
    Buffer Size & $2048-409600$ & how many experiences should be collected before updating the model\\
    Batch Size & $512-5120$ (continuous), $32-512$ (discrete) & number of experiences used for one iteration of a gradient descent update.\\
    Number of Epochs & $3-10$ & number of passes through the eperience buffer during gradient descent\\
    Learning Rate & $1e-5-1e-3$ & strength of each gradient descent update step\\
    Time Horizon & $32-2048$ & number of steps of experience to collect per-agent before adding it to the experience buffer\\
    Max Steps & $5e5-1e7$ & number of steps of the simulation (multiplied by frame-skip) during the training process\\
    Beta & $1e-4-1e-2$ & strength of the entropy regularization, which makes the policy "more random"\\
    Epsilon & $0.1-0.3$ & acceptable threshold of divergence between the old and new policies during gradient descent updating\\
    Normalize & $true/false$ & weather normalization is applied to the vector observation inputs\\
    Number of Layers & $1-3$ & number of hidden layers present after the observation input\\
    Hidden Units & $32-512$ & number of units in each fully connected layer of the neural network\\
    \midrule
    Intrinsic Curiosity Module\\
    \midrule
    Curiosity Encoding Size & $64-256$ & size of hidden layer used to encode the observations within the intrinsic curiosity module\\
    Curiosity Strength & $0.1-0.001$ & magnitude of the intrinsic reward generated by the intrinsic curiosity module\\
    \bottomrule
  \end{tabular}
\end{table}

\newpage
\section{Pseudocode}

PPO-CLIP pseudocode \cite{openai_proximal_2021, schulman_proximal_2017}:

\begin{algorithm}
	\caption{PPO-Clip}
	
	\begin{algorithmic}[1]
		\item Input: initial policy parameters $\theta_0$, initial value function parameters $\phi_0$
		\For {$k=0,1,2,\ldots$}
		    \State Collect set of trajectories $\mathcal{D}_k$ = \{$\tau_i$\} by running policy $\pi_k = \pi(\theta_k)$ in the environment.
		    \State Compute rewards-to-go $\hat{R_t}$.
		    \State Compute advantage estimates, $\hat{A_t}$ (using any method of     advantage estimation) based on the
		    \State current value function $V_{\phi_k}$
			\State Update the policy by maximizing the PPO-Clip objective:
			\State $\theta_{k+1} = arg\underset{\theta}{max} \frac{1}{|\mathcal{D}_k|T} \sum_{\tau \in \mathcal{D}_k} \sum_{t = 0}^{T} \min \left( \frac{\pi_\theta(a_t|s_t)}{\pi_{\theta_k}(a_t|s_t)}A^{\pi_{\theta_k}}(s_t, a_t), g(\epsilon, A^{\pi_{\theta_k}}(s_t, a_t)) \right)$,
			\State typically via stochastic gradient ascent with Adam.
			\State Fit value function by regression on mean-squared error:
			\State $\phi_{k+1} = arg\underset{\phi}{min} \frac{1}{|\mathcal{D}_k|T} \sum_{\tau \in \mathcal{D}_k} \sum_{t = 0}^{T} \left( (V_{\phi}(s_t)-\hat{R_t} \right)$
			\State typically via some gradient descent algorithm.
		\EndFor
	\end{algorithmic} 
\end{algorithm}

Simple Multi-Agent PPO pseudocode:

\begin{algorithm}
	\caption{Multi-Agent PPO} 
	\begin{algorithmic}[1]
		\For {$iteration=1,2,\ldots$}
			\For {$actor=1,2,\ldots,N$}
				\State Run policy $\pi_{\theta_{old}}$ in environment for $T$ time steps
				\State Compute advantage estimates $\hat{A}_{1},\ldots,\hat{A}_{T}$
			\EndFor
			\State Optimize surrogate $L$ wrt. $\theta$, with $K$ epochs and minibatch size $M\leq NT$
			\State $\theta_{old}\leftarrow\theta$
		\EndFor
	\end{algorithmic} 
\end{algorithm}

\newpage
\section{Training}

\begin{figure}[H]
\subsection{Cumulative Reward and Value Loss while Training}
\centering
\includegraphics[width=\linewidth]{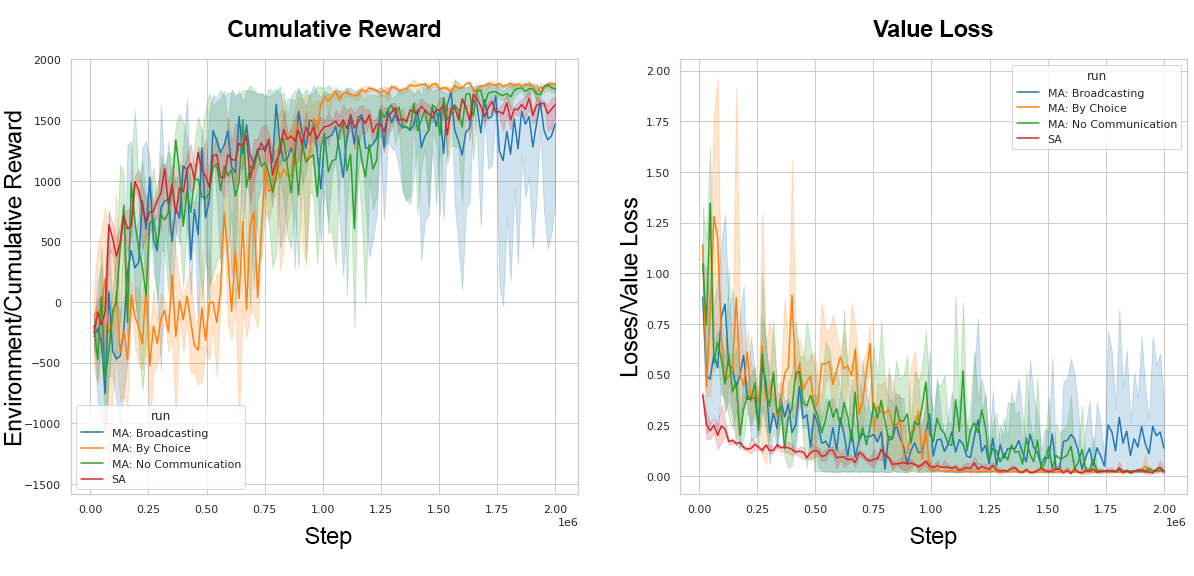}
\caption{Cumulative reward and value loss per time step while training. These diagrams have also been integrated in the main body of the paper: Figure \ref{fig:training-cumulative-reward} and Figure \ref{fig:training-value-loss}.}
\label{fig:cum-reward-value-loss-large}
\end{figure}

\begin{figure}[H]
\subsection{Training and Convergence Time - Scalability}
\centering
\includegraphics[width=\linewidth]{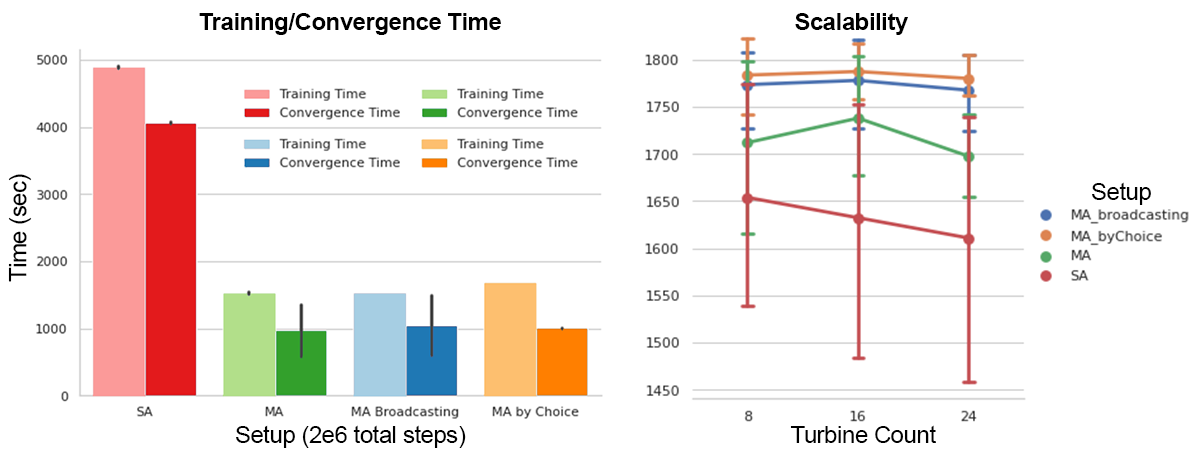}
\caption{Training and convergence time on the left figure. The right figure shows Scalability in inference mode on a random distribution pattern: cumulative reward while scaling from 8 to 16 and 24 turbines per wind farm. These diagrams have also been integrated in the main body of the paper: Figure \ref{fig:training-time-convergence} and Figure \ref{fig:scalability}.}
\label{fig:time-scalability-large}
\end{figure}

\newpage
\section{Inference}
\subsection{20 times 1e6 Inference Mode Steps for All Setups}
\begin{figure}[H]
\centering
\includegraphics[width=\linewidth]{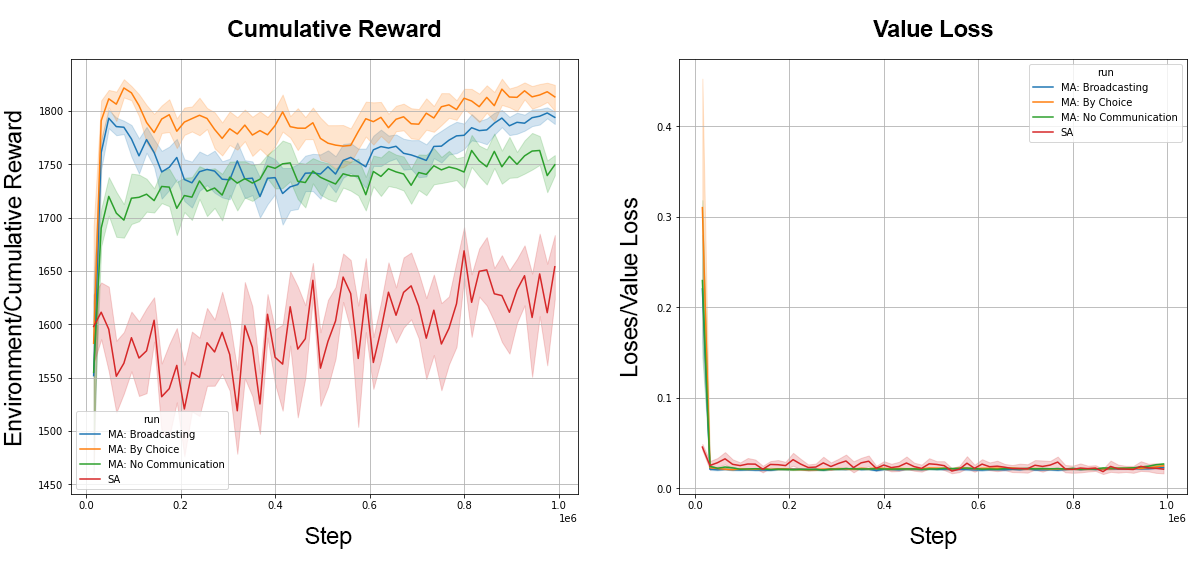}
\caption{Inference mode benchmarking, for each setup, 20 times for 1e6 time steps. Default distribution pattern: Cumulative Reward and Value Loss. These diagrams have also been integrated in the main body of the paper: Figure \ref{fig:cum-reward-inference} and \ref{fig:value-loss-inference}.}
\label{fig:cum-reward-value-loss-inference-large}
\end{figure}

\subsection{Setup VS Total Cumulative Reward}
\begin{figure}[!h]
\centering
\includegraphics[width=0.9\linewidth]{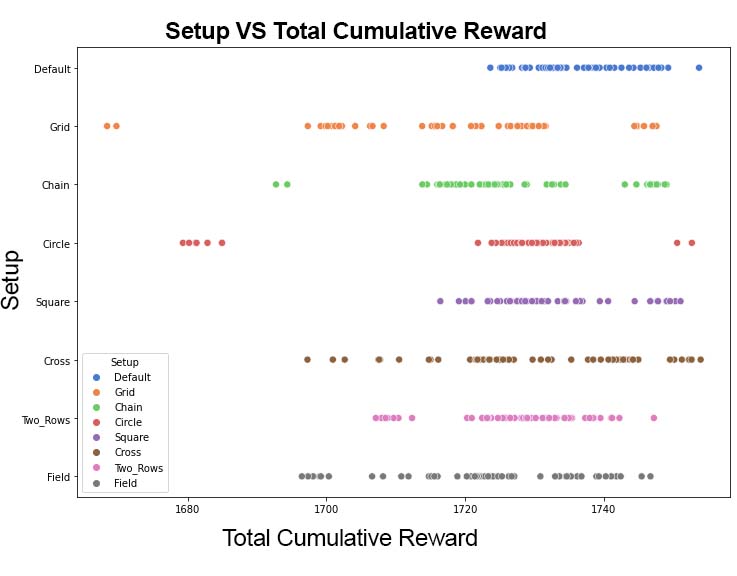}
\caption{The diagram visualises the relationship between the distribution pattern setup and the cumulative reward. MA system seem so achieve highest cumulative rewards on the Default distribution pattern.}
\label{fig:setup-vs-cum-reward}
\end{figure}

\newpage
\subsection{Layout and Communication Graphs: Agent-Setup VS Layout}
\begin{figure}[H]
  \includegraphics[width=\textwidth]{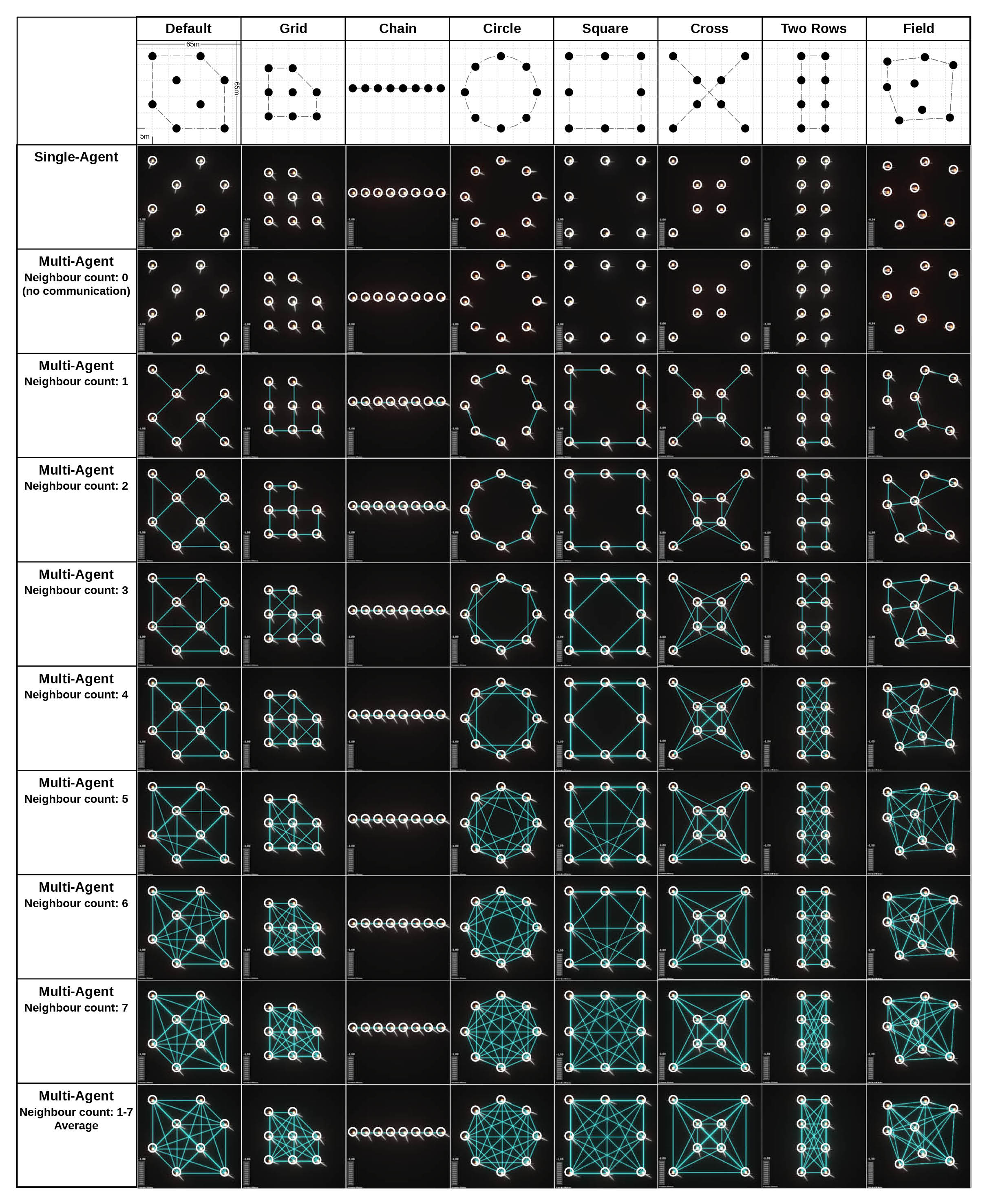}
  \caption{Table to show graphs build by nearest neighbour algorithm for variation of distribution patterns. The graphs are used by the GNN Message Passing communication layer in the communication setup case.}
\end{figure}

\newpage
\subsection{Turbine Performance MA Broadcasting in Inference Mode: Agent-Setup VS Layout}
\begin{figure}[H]
  \includegraphics[width=\textwidth]{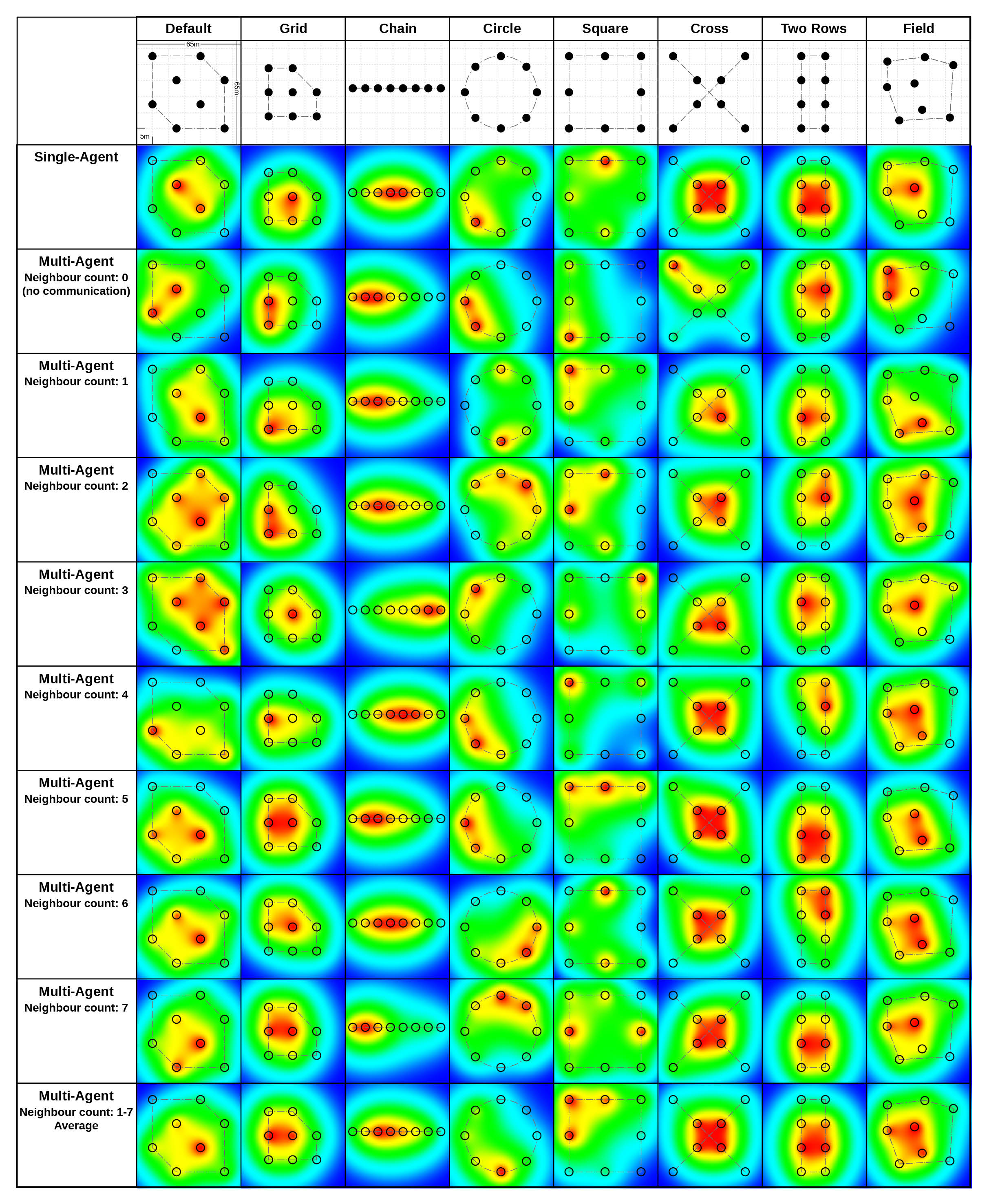}
  \caption{Individual performance mapping for distribution patterns using the MA Broadcasting setup. We mapped the individual performance of agents and turbines. Red being high and blue low cumulative reward. It is fairly easy to see that agent and turbines surrounded by others perform better than located on the perimeter.}
\end{figure}

\newpage
\subsection{Turbine Performance MA by Choice in Inference Mode: Agent-Setup VS Layout}
\begin{figure}[H]
  \includegraphics[width=\textwidth]{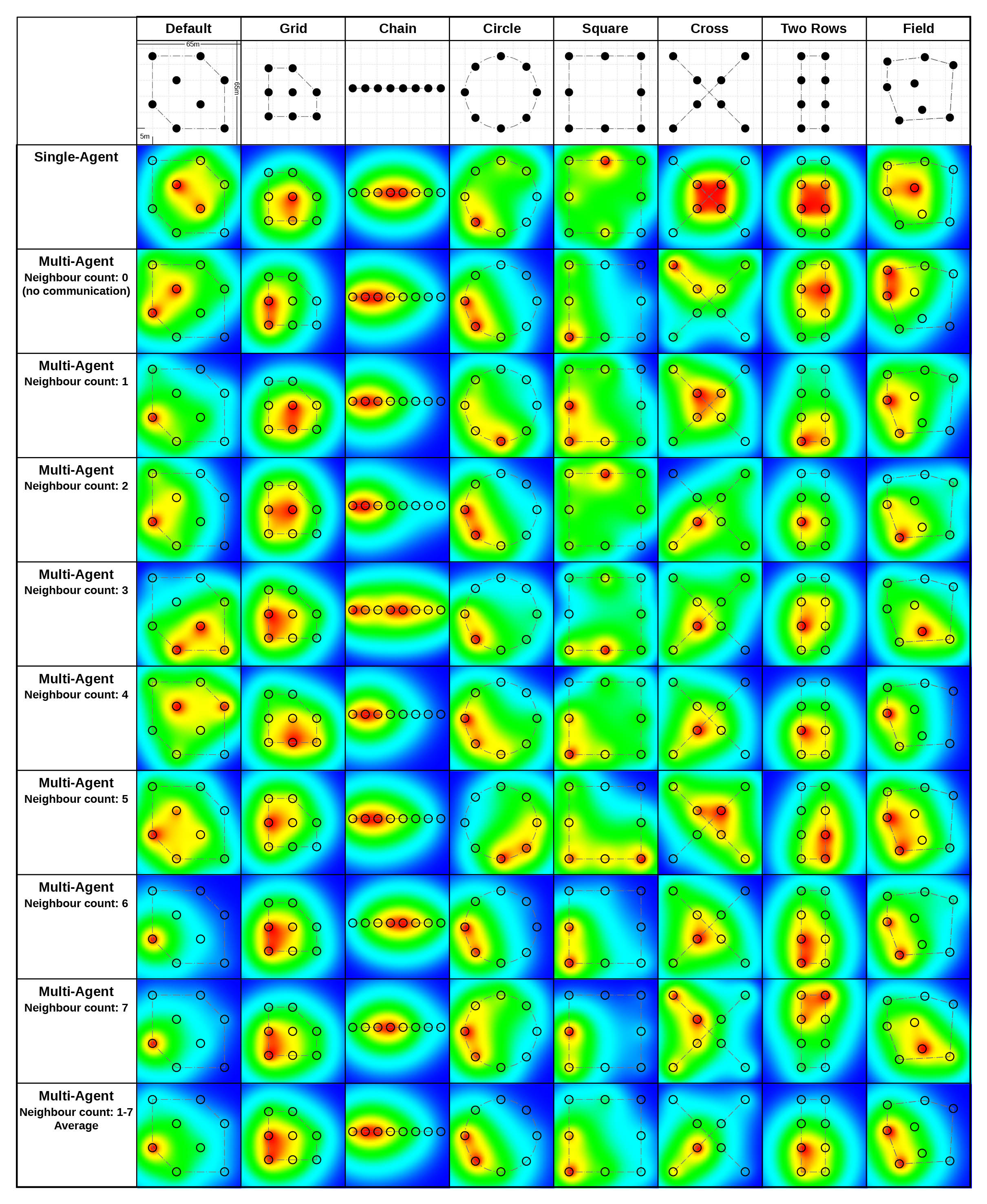}
  \caption{Individual performance mapping for distribution patterns using the MA by Choice setup. We mapped the individual performance of agents and turbines. Red being high and blue low cumulative reward. It is fairly easy to see that agent and turbines surrounded by others perform better than located on the perimeter.}
\end{figure}

\newpage
\section{Communication}
\subsection{Graph Neural Network Message Passing Communication Layer}
\begin{figure}[!h]
\includegraphics[width=0.9\linewidth]{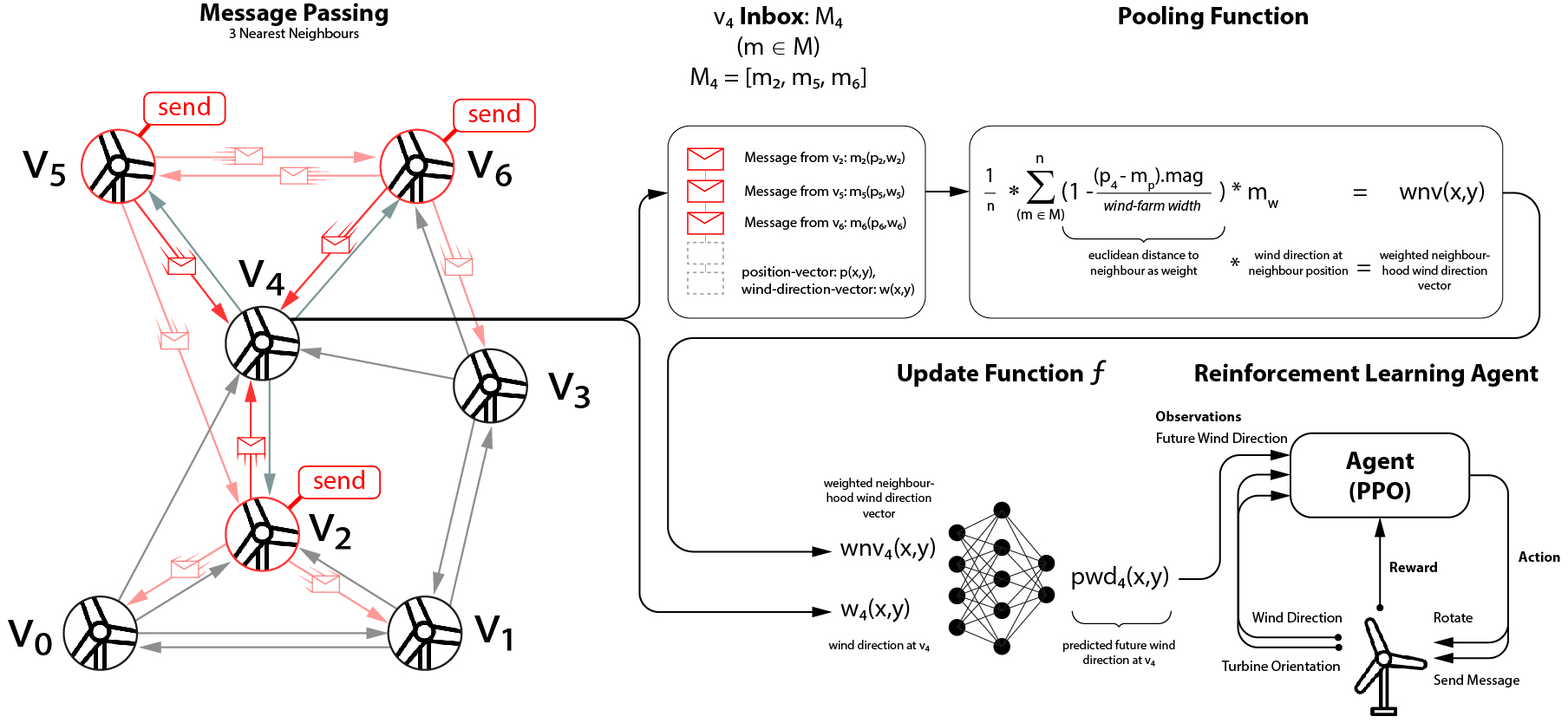} 
\caption{The diagram is a supplement to describe the GNN Message Passing implementation. Messages are received and pooled using euclidean distance weighted neighbourhood wind directions. The resulting vector and current wind direction at the turbine are used as input to train a neural network. The ground truth is the future wind direction at the turbine. This diagram has also been integrated in the main body of the paper: Figure \ref{fig:GNN_MessagePassing}.}
\label{fig:GNN_MessagePassing-large}
\end{figure}

\subsection{Communication Count VS Wind Direction - Turbine Orientation Delta}
\begin{figure}[!h]
\centering
\includegraphics[width=0.85\linewidth]{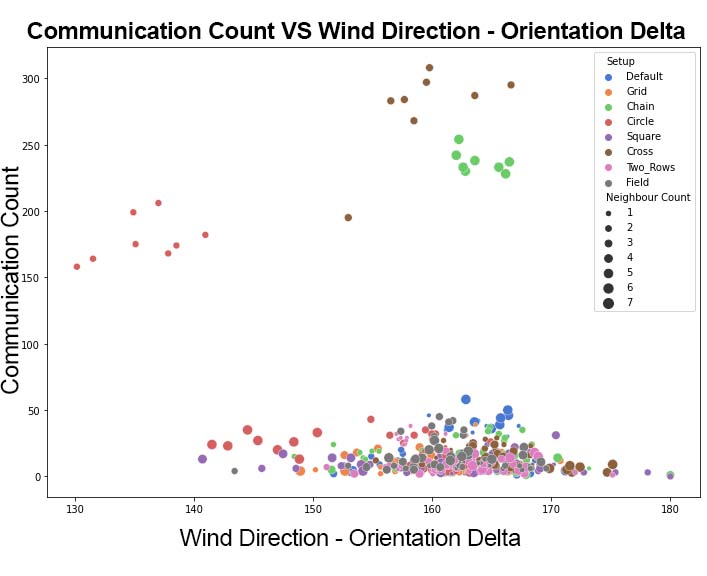}
\caption{The diagram shows the delta between the wind turbine and the wind direction at that turbine in relation to the communication count. More communication is occurring when the delta is high, which also means that the wind turbine performance is high. This diagram has also been integrated in the main body of the paper: Figure \ref{fig:com-count-vs-wind-dir-orientation}.}
\label{fig:com-count-vs-wind-dir-orientation-large}
\end{figure}

\subsection{Wind Direction at Turbine when Communicating (sending messages)}
\begin{figure}[!h]
\centering
\includegraphics[width=0.85\linewidth]{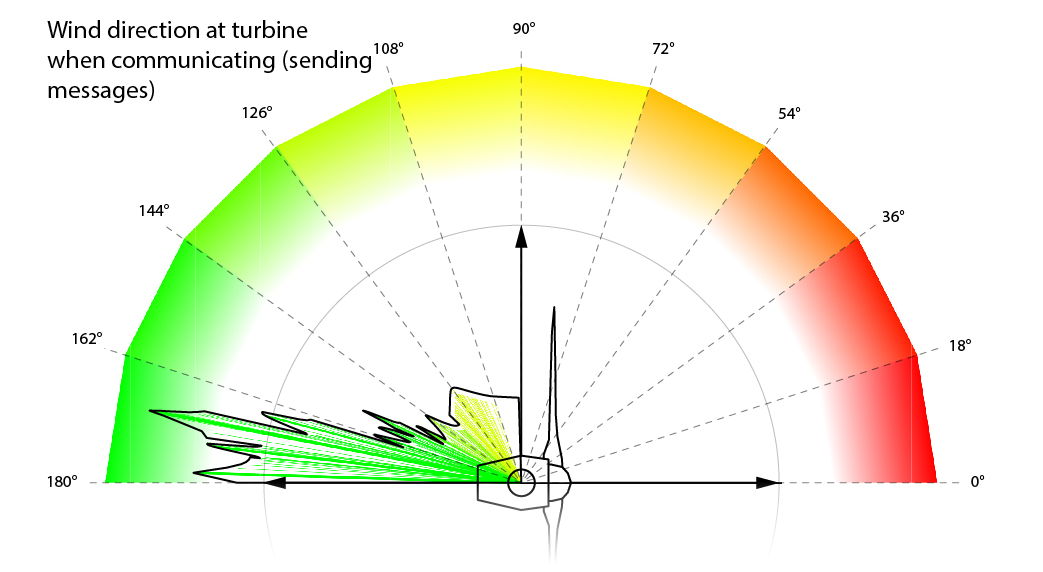}
\caption{This diagram shows the wind direction at the wind turbine, at the time step of communication. The length of radial lines indicate how often communication at that angle occurred. The information we can read from this diagram is that the turbine is more likely to send a message when performing well itself. This diagram has also been integrated in the main body of the paper: Figure \ref{fig:wind-dir-at-turbine-when-com}.}
\label{fig:wind-dir-at-turbine-when-com-large}
\end{figure}

\subsection{Communication Count VS Total Cumulative Reward}
\begin{figure}[!h]
\centering
\includegraphics[width=0.9\linewidth]{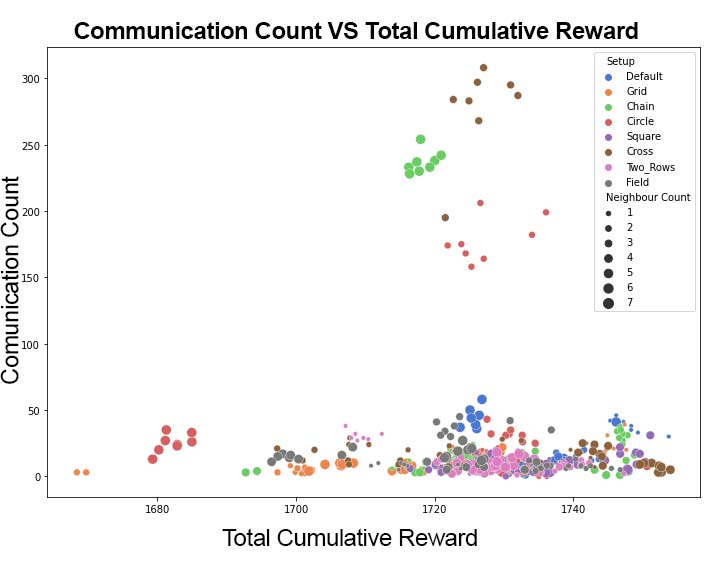}
\caption{This diagram shows that high a communication count results in high cumulative rewards.This diagram has also been integrated in the main body of the paper: Figure \ref{fig:com-count-vs-cum-reward}.}
\label{fig:com-count-vs-cum-reward-large}
\end{figure}

\newpage
\section{Nearest Neighbour Count Benchmarking}
\subsection{Neighbour Count VS Total Cumulative Reward}
\begin{figure}[H]
\centering
\includegraphics[width=\linewidth]{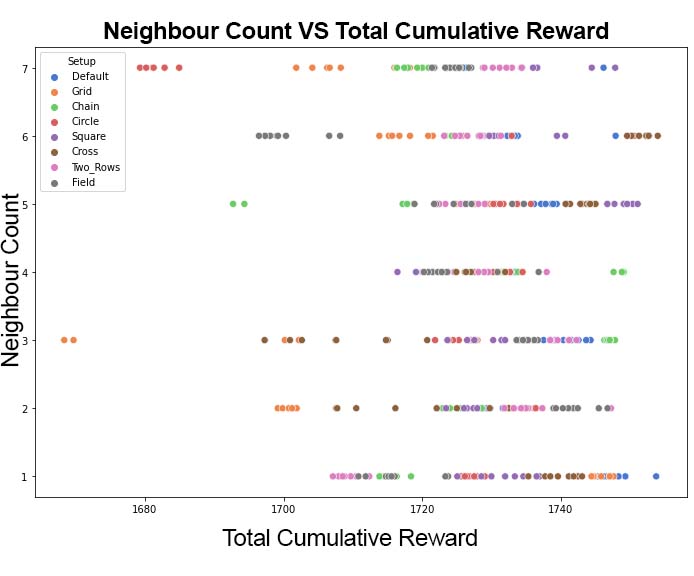}
\caption{Comparison between the neighbour count and the resulting cumulative reward. Color of the dots indicate the distribution pattern. Neighbour count of 4 seems to be a preferable neighbour count. This diagram has also been integrated in the main body of the paper: Figure \ref{fig:n-count-vs-cum-reward}.}
\label{fig:n-count-cum-reward-large}
\end{figure}

\newpage
\subsection{Turbine Communication Count for MA by Choice in Inference Mode: Agent-Setup VS Layout}
\begin{figure}[H]
\centering
  \includegraphics[width=\textwidth]{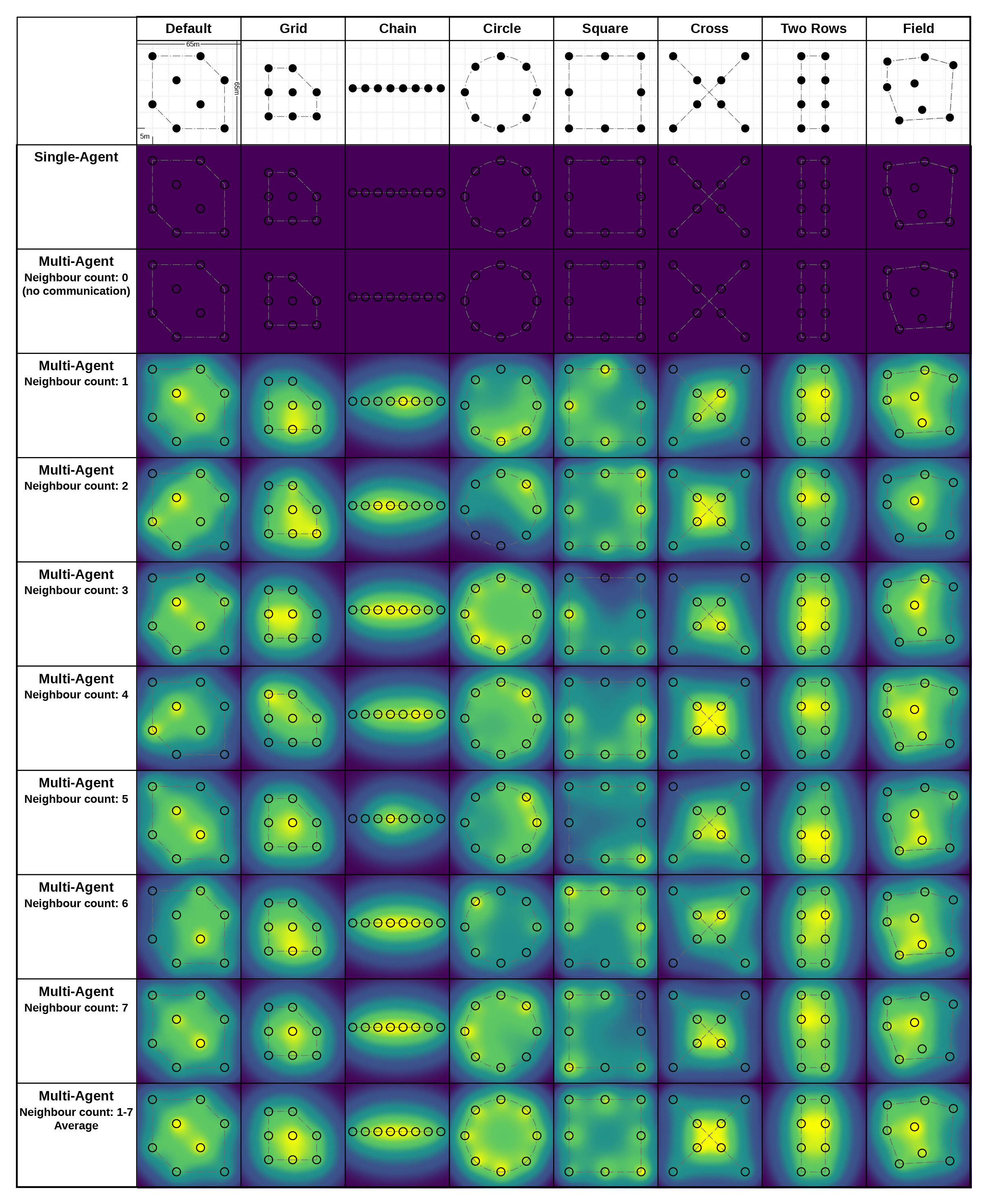}
  \caption{Individual communication mapping for discrete distribution patterns. We mapped where communication is occurring the most. In parallel with the performance mapping, we can see that high performing agents communicate the most or vice versa agents that communicate most yield the highest performance.}
\end{figure}

\newpage
\subsection{Multi-Agent Broadcasting - Default Layout - 0 Nearest Neighbours (no communication) - 10 runs each 1e6 steps (2000 each episode)}
\begin{figure}[H]
\centering
  \includegraphics[width=0.8\textwidth]{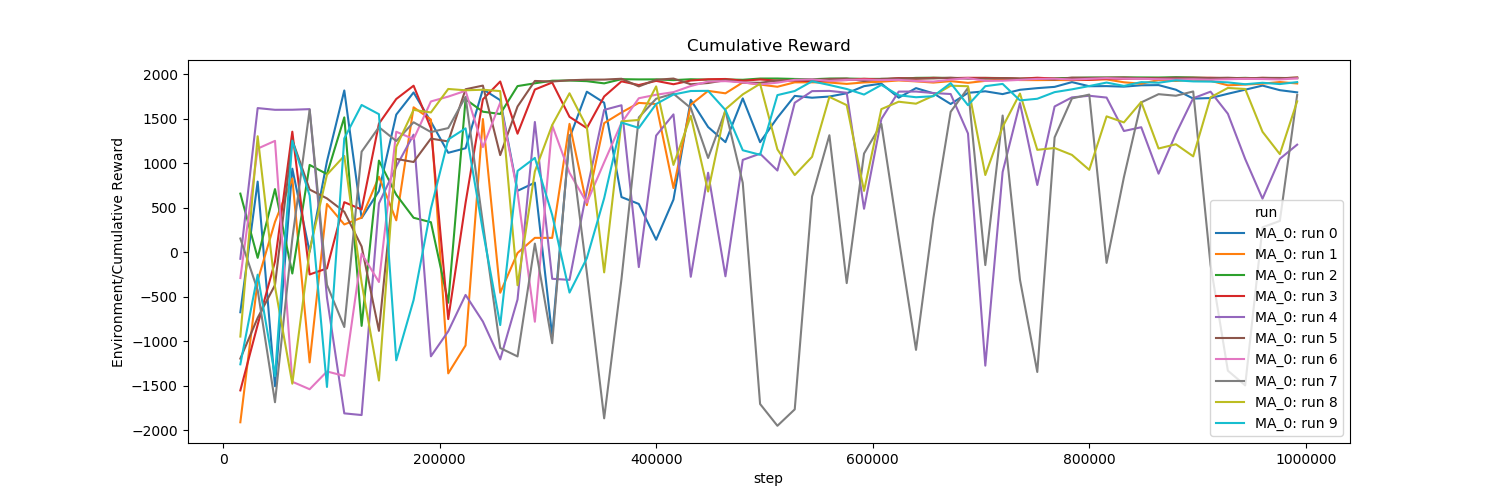}
  \caption{Cumulative Reward; 1e6 total time steps.}
\end{figure}
\begin{figure}[H]
\centering
  \includegraphics[width=0.8\textwidth]{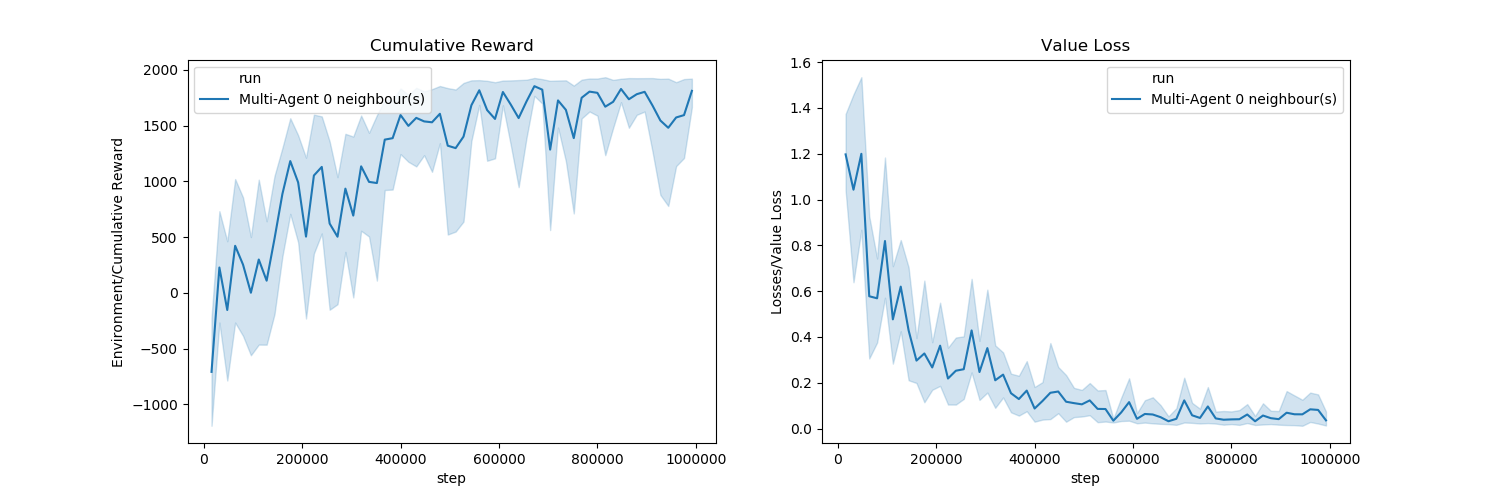}
  \caption{Mean cumulative reward and value loss.}
\end{figure}

\subsection{Multi-Agent - Default Layout - 1 Nearest Neighbour(s) - 10 runs each 1e6 steps (2000 each episode)}
\begin{figure}[H]
\centering
  \includegraphics[width=0.8\textwidth]{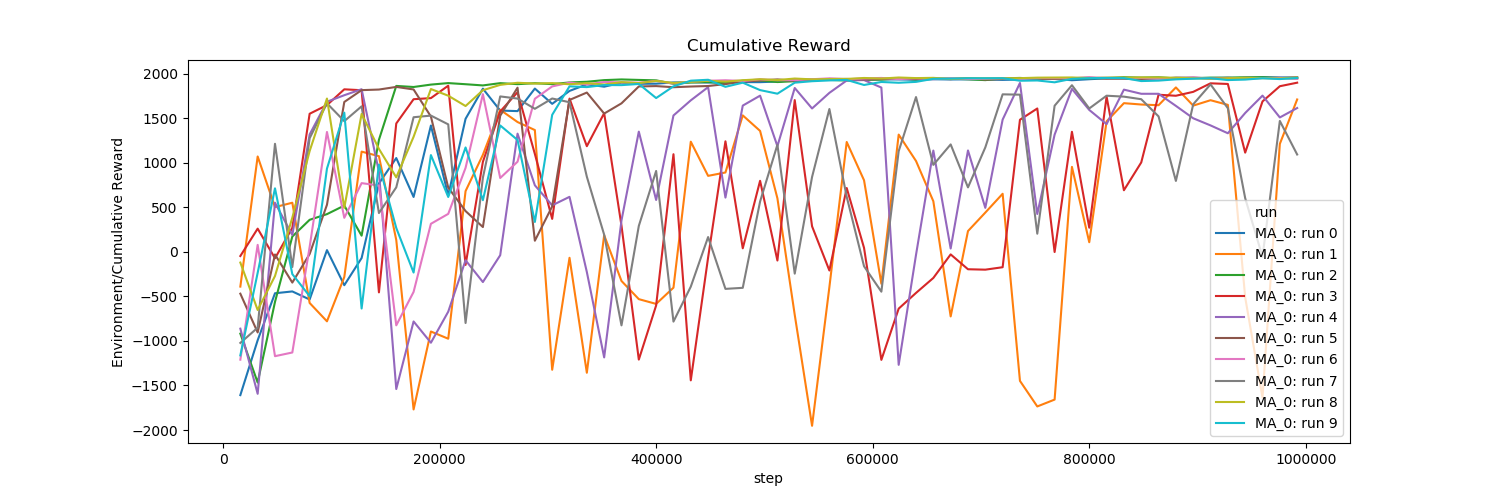}
  \caption{Cumulative Reward; 1e6 total time steps.}
\end{figure}
\begin{figure}[H]
\centering
  \includegraphics[width=0.8\textwidth]{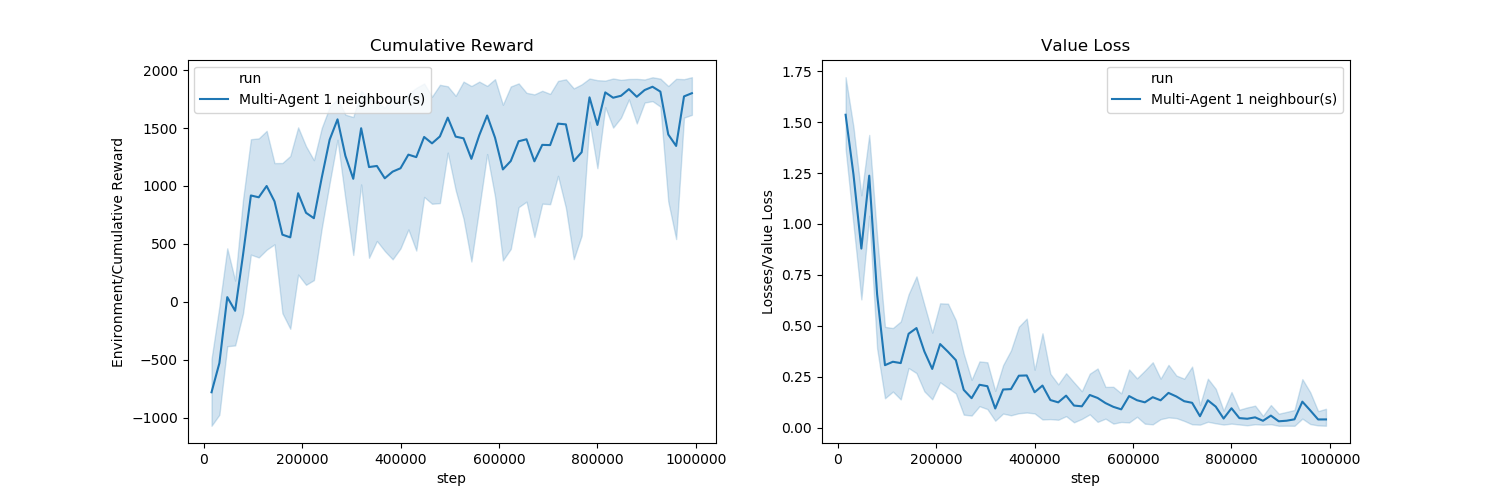}
  \caption{Mean cumulative reward and value loss.}
\end{figure}

\newpage
\subsection{Multi-Agent - Default Layout - 3 Nearest Neighbour(s) - 10 runs each 1e6 steps (2000 each episode)}
\begin{figure}[H]
\centering
  \includegraphics[width=0.8\textwidth]{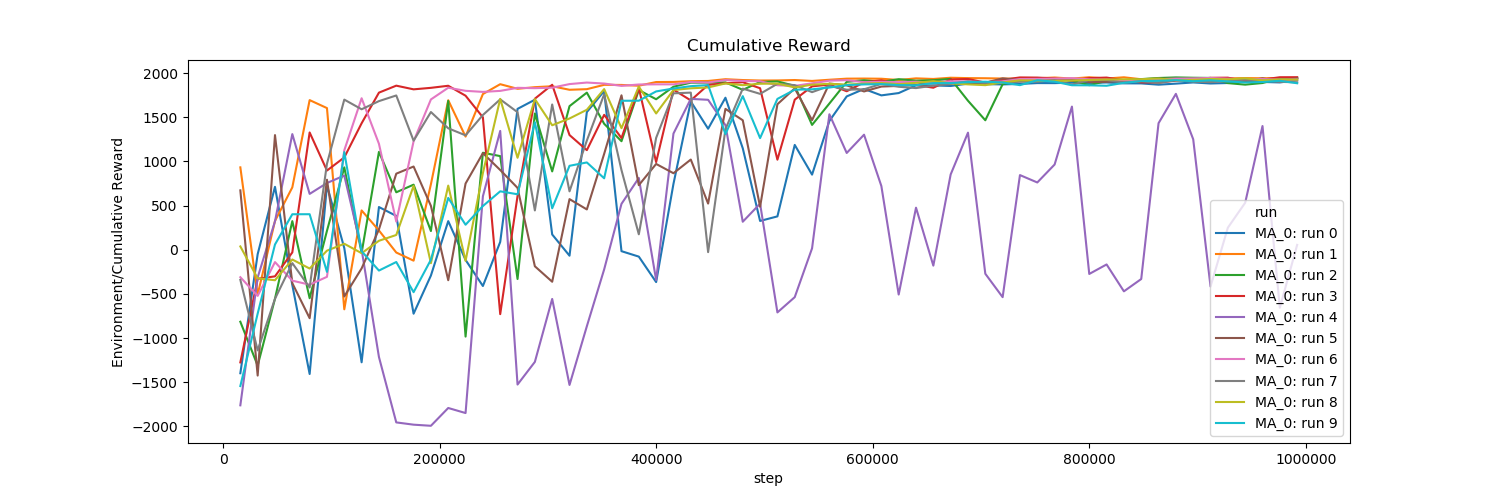}
  \caption{Cumulative Reward; 1e6 total time steps.}
\end{figure}
\begin{figure}[H]
\centering
  \includegraphics[width=0.8\textwidth]{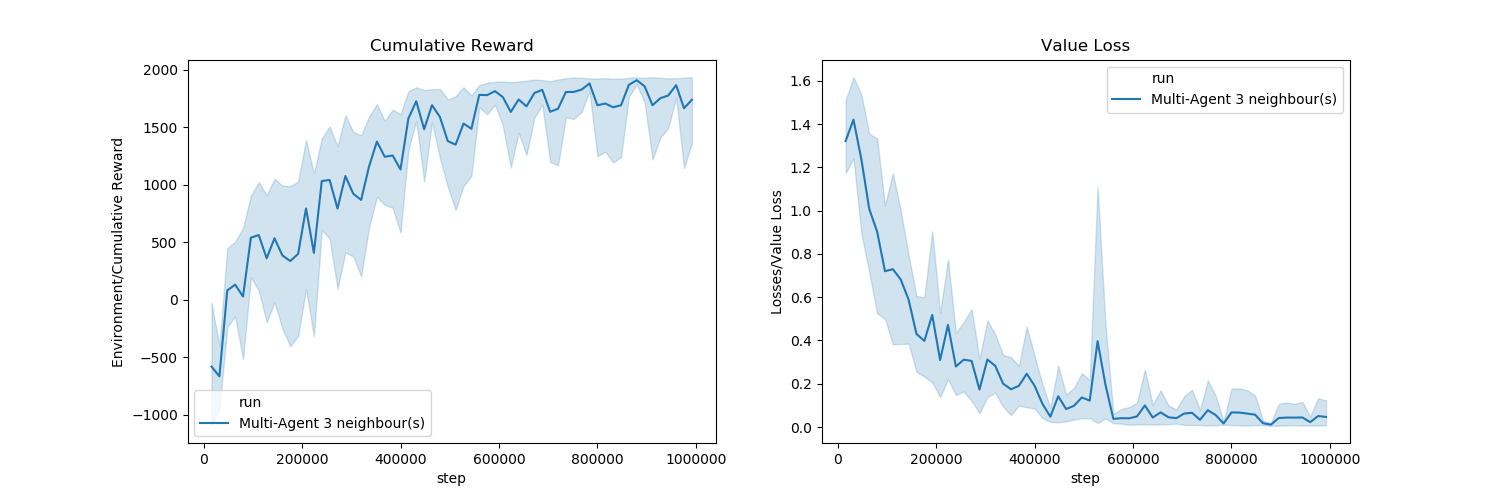}
  \caption{Mean cumulative reward and value loss.}
\end{figure}

\subsection{Multi-Agent - Default Layout - 5 Nearest Neighbour(s) - 10 runs each 1e6 steps (2000 each episode)}
\begin{figure}[H]
\centering
  \includegraphics[width=0.8\textwidth]{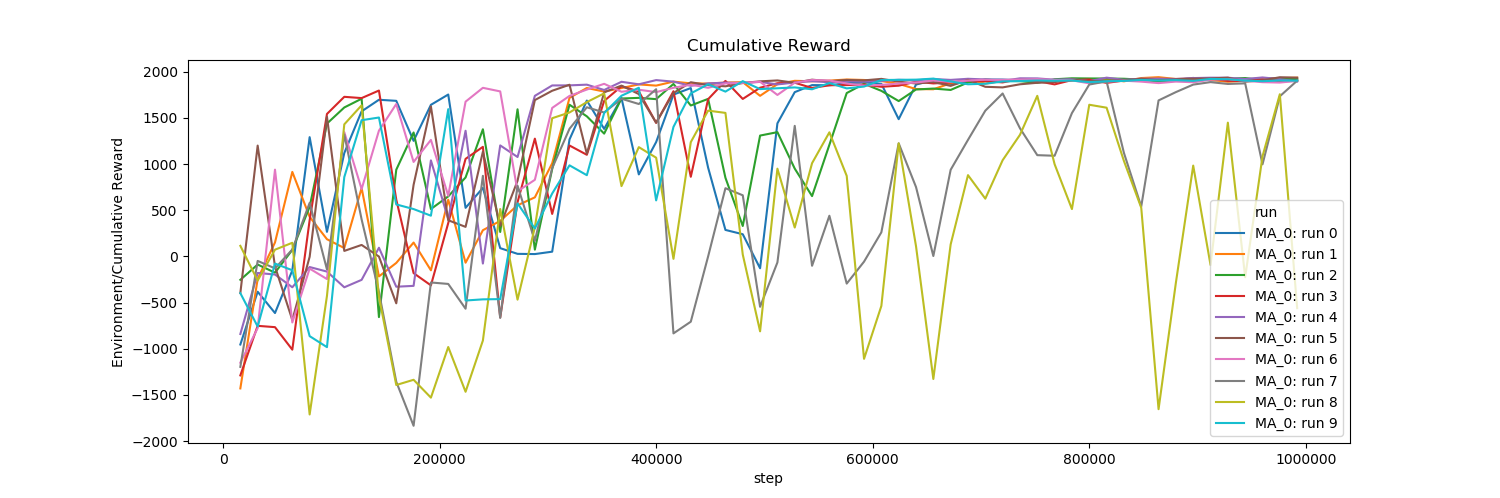}
  \caption{Cumulative Reward; 1e6 total time steps.}
\end{figure}
\begin{figure}[H]
\centering
  \includegraphics[width=0.8\textwidth]{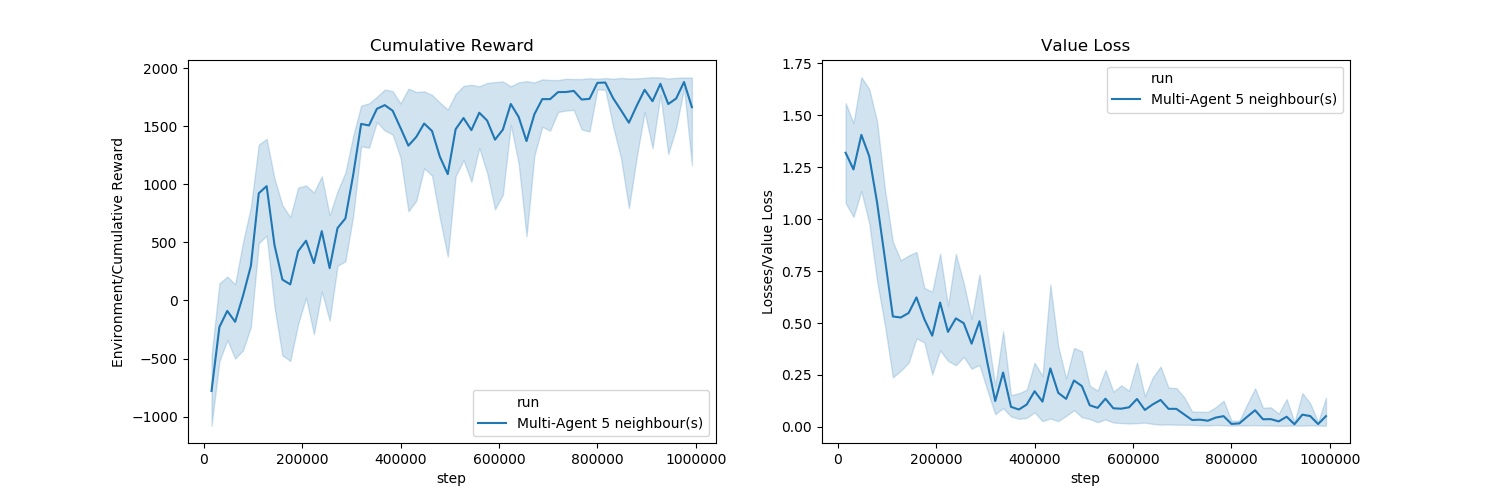}
  \caption{Mean cumulative reward and value loss.}
\end{figure}

\newpage
\section{Additional Environment Screenshots}
\subsection{Additional Environment Screenshots: Inference Mode}
\begin{figure}[H]
  \includegraphics[width=\textwidth]{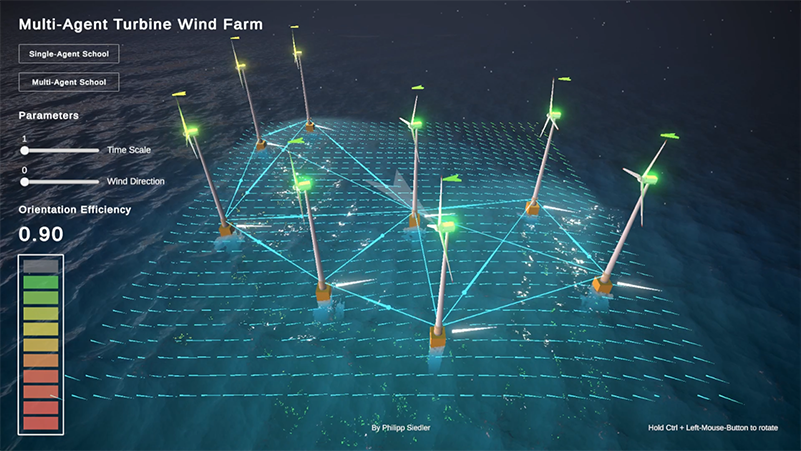}
  \caption{Sample Random distribution environment in inference mode.}
\end{figure}
\begin{figure}[H]
  \includegraphics[width=\textwidth]{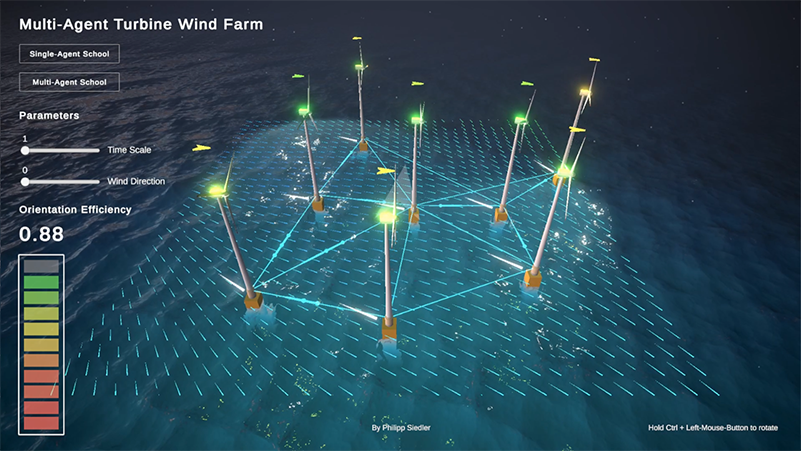}
  \caption{Sample Random distribution environment in inference mode.}
\end{figure}

\newpage
\subsection{Additional Environment Screenshots: Training School}
\begin{figure}[H]
  \includegraphics[width=\textwidth]{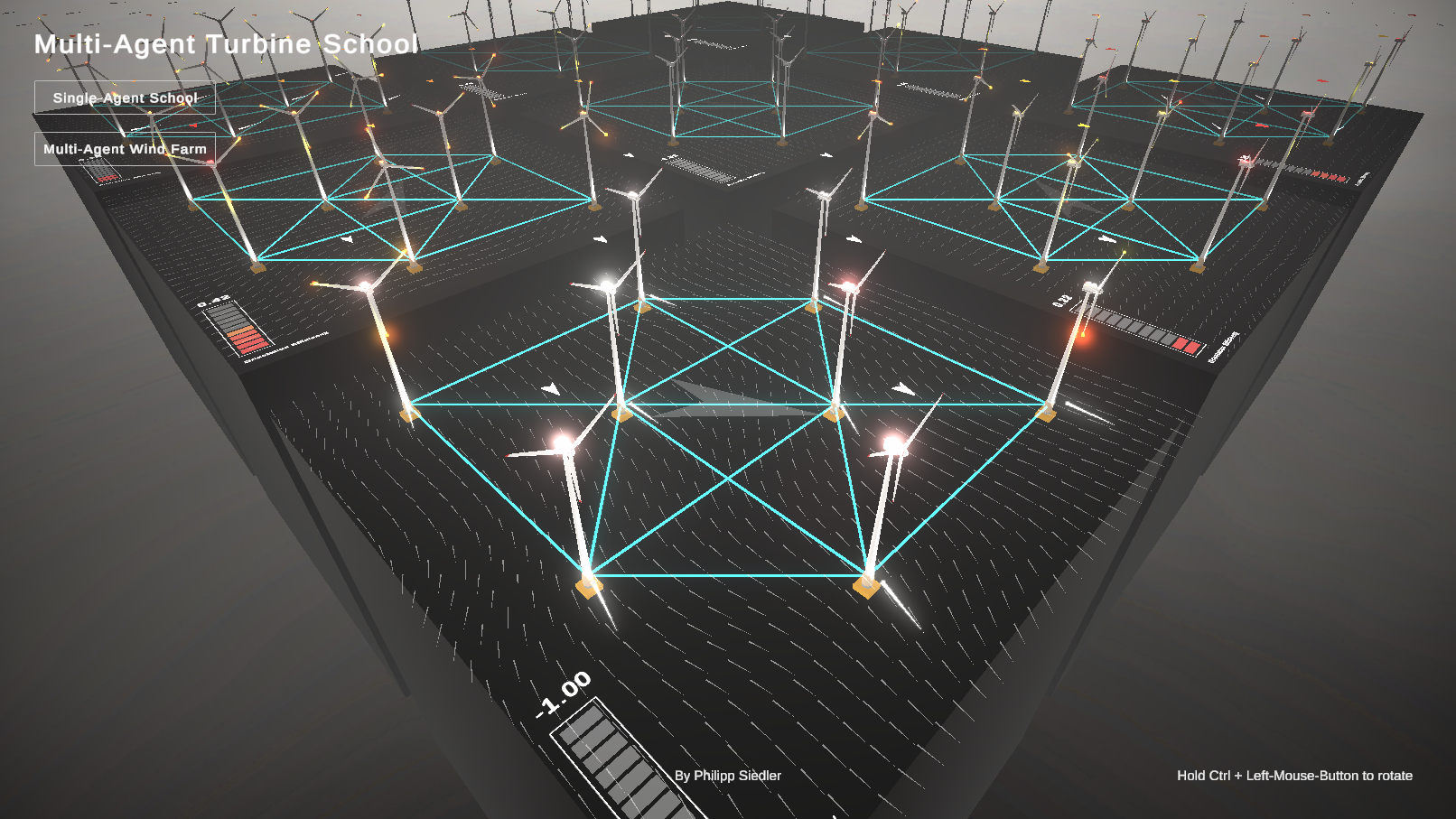}
  \caption{9 parallel sample Default distribution pattern training environments for setup: MA by Choice and Broadcasting.}
\end{figure}
\begin{figure}[H]
  \includegraphics[width=\textwidth]{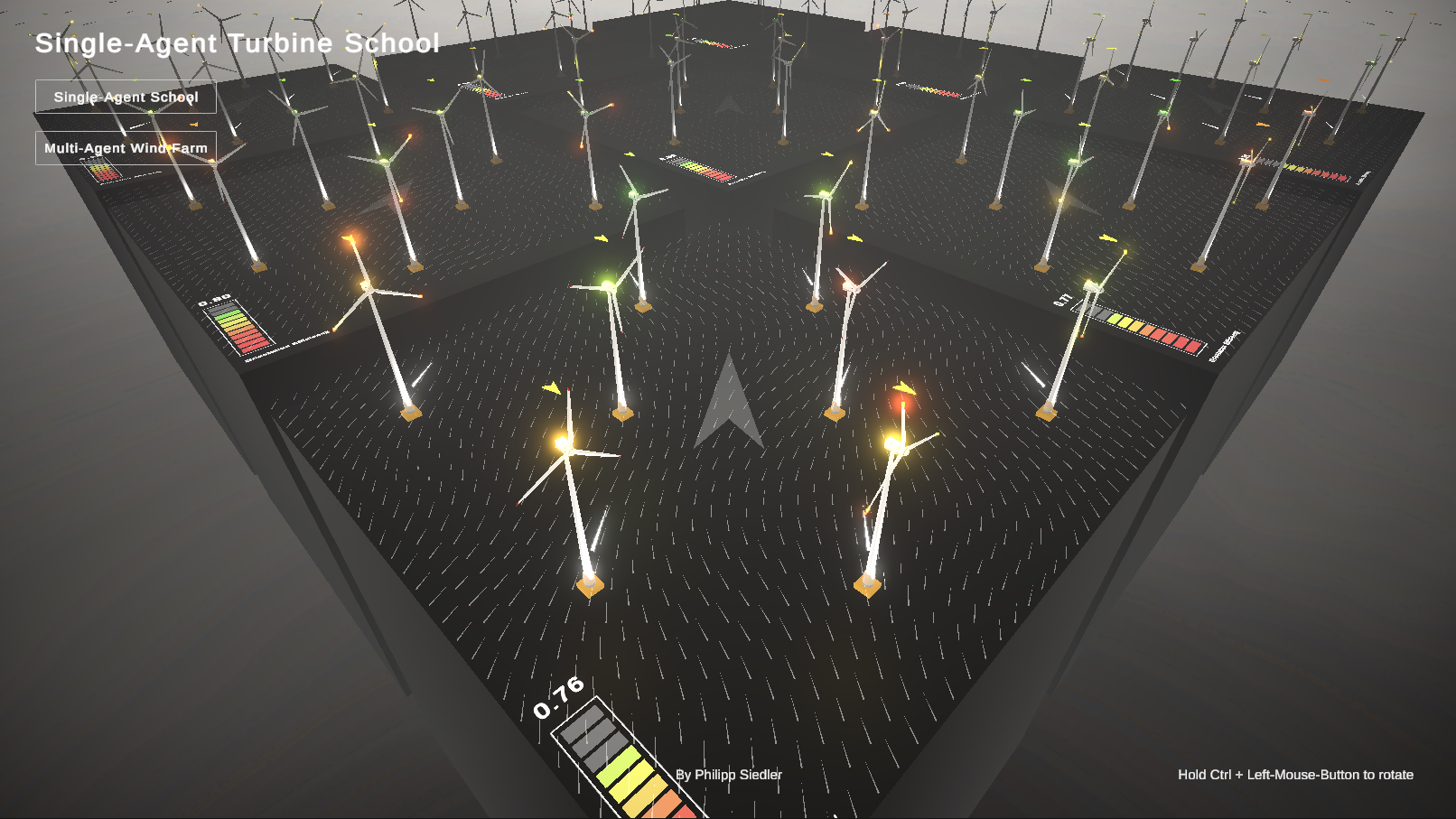}
  \caption{9 parallel sample Default distribution pattern training environments for setup: SA and MA no Communication.}
\end{figure}
\end{document}